**Title:**

# RL-100: Performant Robotic Manipulation with Real-World Reinforcement Learning

**Authors:**

Kun Lei[1,2†#], Huanyu Li[1,2†], Dongjie Yu[1,3†], Zhenyu Wei[5†], Lingxiao Guo[6], Zhennan Jiang[7], Ziyu Wang[4], Shiyu Liang[2], Huazhe Xu[1,4*]

**Affiliations:**

[1]Shanghai Qi Zhi Institute; Shanghai, China
[2]School of Computer Science, Shanghai Jiao Tong University; Shanghai, China
[3]School of Computing and Data Science, The University of Hong Kong; HKSAR, China
[4]Institute for Interdisciplinary Information Sciences, Tsinghua University; Beijing, China
[5]Department of Computer Science, University of North Carolina at Chapel Hill; Chapel Hill, USA
[6]School of Computer Science, Carnegie Mellon University; Pittsburgh, USA
[7]Institute of Automation, Chinese Academy of Sciences; Beijing, China
[*]Corresponding author. Email: huazhe_xu@mail.tsinghua.edu.cn
[†]These authors contributed equally to this work.
[#] This author is the project lead for this work.

**Abstract:**

Real-world robotic manipulation in homes and factories demands reliability, efficiency, and robustness that approach or surpass skilled human operators. We present RL-100, a real-world reinforcement learning framework built on diffusion visuomotor policies. RL-100 unifies imitation and reinforcement learning under a single clipped PPO surrogate objective applied within the denoising process, yielding conservative, stable improvements across offline and online stages. To meet deployment latency, a lightweight consistency distillation compresses multi-step diffusion into a one-step controller for high-frequency control. The framework is task-, embodiment-, and representation-agnostic, and supports both single-action and action-chunking control. We evaluate RL-100 on eight diverse real-robot tasks—from dynamic pushing and agile bowling to pouring, cloth folding, unscrewing, multi-stage juicing, long-horizon box folding. RL-100 attains 100% success across evaluated trials for a total of 1000 out of 1000 episodes, including up to 250 out of 250 consecutive trials on one task. It matches or surpasses expert teleoperators in time-to-completion. Without retraining, a single policy attains approximately 90% zero-shot success under environmental and dynamics shifts, adapts in a few-shot regime to significant task variations (86.7%), and remains robust to aggressive human perturbations (about 96%). Notably, our juicing robot served random customers continuously for about seven hours without failure when zero-shot deployed in a shopping mall. These results suggest a practical path to deployment-ready robot learning by starting from human priors, aligning training objectives with human-grounded metrics, and reliably extending performance beyond human demonstrations. For more results and videos, please visit our anonymous project website: https://lei-kun.github.io/RL-100/

**One-Sentence Summary:**

RL-100 attains 100% success on eight robot tasks with human-level efficiency and seven-hour mall deployment without retraining.

**Main Text:**

# INTRODUCTION

Dexterous robotic manipulation stands as an iconic challenge in robotics (*1*, *2*). Real-world deployment beyond laboratories requires human-level reliability, efficiency, and robustness. Recent learning-based advances across generative diffusion policies (*3*, *4*), diffusion-based robot foundation models (*5*, *6*), sim-to-real RL (*7*, *8*), and real-world RL (*2*, *9*) have narrowed this gap, demonstrating human-like manipulation proficiency. In particular, generative policies and foundation models are trained or fine-tuned on high-quality, human-collected real-robot datasets at varying scales, providing strong human priors and enabling robots to acquire the efficient strategies used by skilled tele-operators. However, high-quality real-robot data remain scarce: teleoperation incurs perceptual and control latency and favors slow, conservative motions (*10*). Moreover, large-scale collection depends on skilled operators and is labor-intensive and expensive. As a result, state–action coverage is limited, undermining generalization and reliability. Consequently, this supervised paradigm is constrained by an imitation ceiling: under purely supervised objectives, performance is effectively bounded by demonstrator skill and inherits human inefficiencies, biases, and occasional errors.

Reinforcement learning (RL) offers a complementary route by optimizing returns from interaction rather than imitation fidelity, enabling the discovery of strategies that are rare or absent in human demonstrations. At the same time, sim-to-real RL must contend with visual and dynamics gaps between simulation and reality, while naïvely training a learning-based generative policy on real hardware is risky and sample-inefficient. This raises a central question: how can we build a robotic learning system that leverages strong human priors yet continues to refine itself through autonomous exploration? A useful analogy comes from human learning: babies learn to walk under parental supervision, then reinforce the skill on their own until they master it and eventually transfer it across terrains. Analogously, a generalizable robotic learning system should combine skilled human priors with self-improvement to reach—and in some cases exceed—human capability in reliability, efficiency, and robustness.

In this paper, we introduce RL-100, a framework that employs a real-world RL post-training phase on top of an imitation-based diffusion policy, preserving its expressive strengths while explicitly optimizing deployment metrics – success rate, time-to-completion, and robustness – under mild human-guided exploration. In short, we **start from human** priors, **align with human**-grounded objectives, and **go beyond human** performance. RL-100 has three stages with distinct roles and costs: (i) imitation learning (IL) pretraining on teleoperated demonstrations produces a competent, low-variance policy that anchors behavior in safe, human-like actions; (ii) iterative offline RL post-training, i.e., offline updates on a growing buffer of policy rollouts, delivers the bulk of improvement in success and efficiency across iterations; and (iii) brief on-policy online RL supplies last-mile reliability by targeting rare failure modes that remain after offline refinement, but is the most resource-intensive stage on real hardware (parameter tuning, resets, and approvals). We therefore allocate most of the learning budget to iterative offline updates and use a small, targeted online budget to push performance from high success (e.g., 90%) to near-perfect reliability (e.g., 99%+).

Moreover, RL-100 is representation-agnostic: it operates in a vision-based setting and supports both 3D point clouds and 2D RGB images by simply swapping observation encoders, without modifying rest of the framework. While our real-robot experiments use 3D point clouds as the primary representation, ablations in simulation show the same performance trends with 2D inputs. In particular, we introduce a self-supervised visual encoder tailored for RL post-training, which furnishes stable, task-relevant features throughout policy exploration and updates.

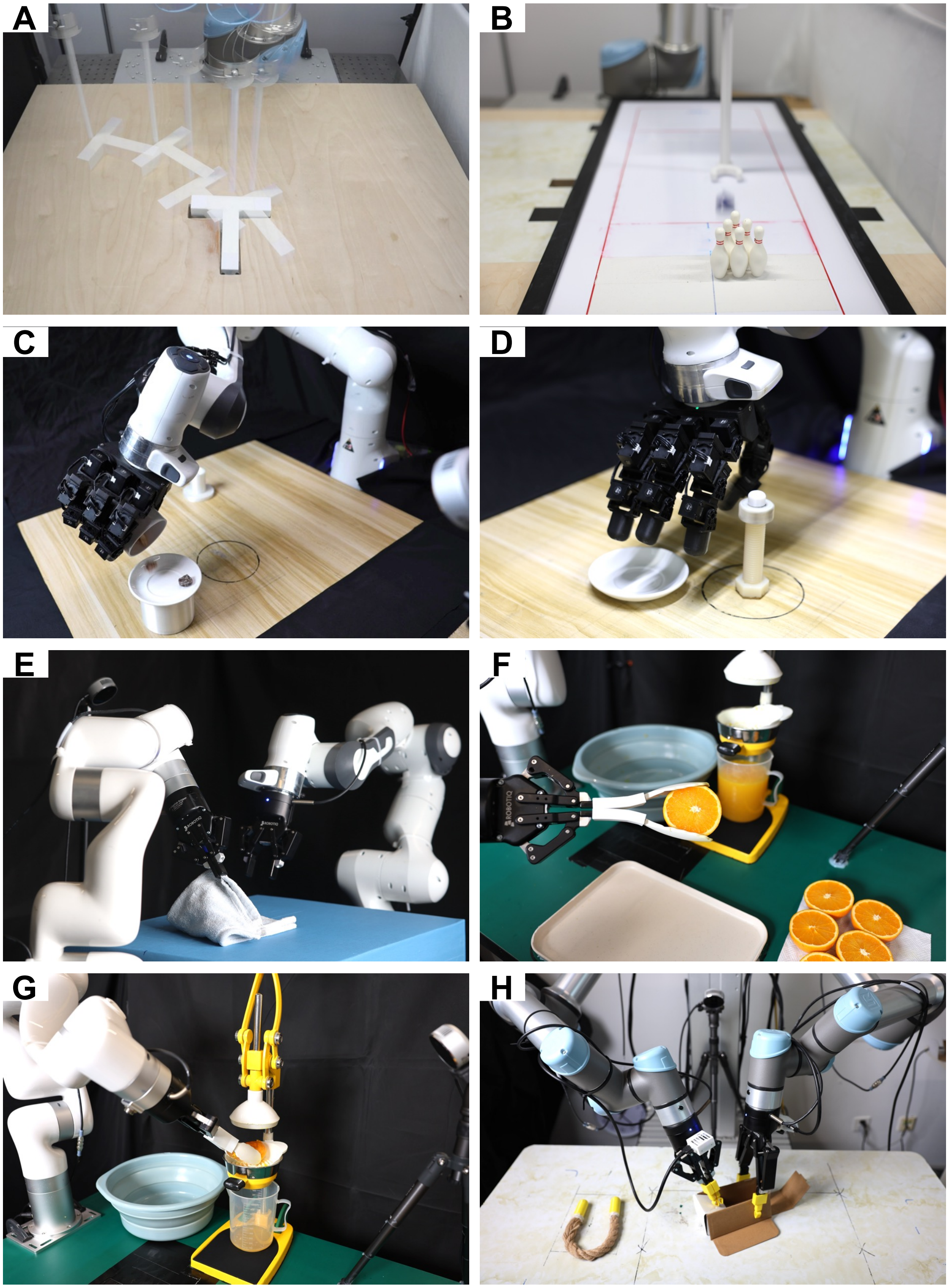

**Fig. 1. Real-robot snapshots illustrating the diversity of our task suite**. (A) Dynamic Push-T; (B) Agile Bowling; (C) Pouring. (D) Dynamic Unscrewing with a dexterous hand; (E) Dual-arm Folding; (F) Orange Juicing - Placing, (G) Orange Juicing - Removal; and (H) Box folding are shown side-by-side in the same panel.

During policy rollouts (for all tasks except Dynamic Push-T), a human operator assigns a **terminal** success reward of +1 and ends the episode when the task is successfully completed; otherwise, the reward remains 0, and episodes terminate by timeout (or early stop for safety) with reward 0. We use a unified policy-gradient objective across both iterative offline and online phases to fine-tune the diffusion sampler's short-horizon denoising schedule (*11*). This alignment yields stable updates across phases and strong fine-tuning sample efficiency. In addition, we interleave a lightweight distillation loss that compresses the $K$-step diffusion policy into a one-step consistency (*12*) policy for deployment, reducing inference latency while maintaining or improving efficiency and robustness.

Moreover, our framework is task- and embodiment-agnostic. We evaluate RL-100 across simulated tasks and on a real-world suite of eight manipulation tasks as illustrated in Fig. 1 and summarized in Table S1: Dynamic Push-T, Agile Bowling, Pouring, Soft-towel Folding, Dynamic Unscrewing, Orange Juicing – Placing, Orange Juicing – Removal, and Box Folding. Orange Juicing comprises two subtasks, placing and removal, which are trained and evaluated separately but reported as one task family. The suite includes rigid-body dynamics, deformable objects, fluids, and precision assembly, and the framework is deployed across multiple embodiments. In particular, Box Folding emphasizes long-horizon bimanual coordination with error-sensitive crease/edge alignment and complex contact dynamics, making it a stringent test of robustness, adaptation, and spatial generalization. For these tasks and embodiments, we select a specific control mode per task: a single-step control mode is used when a fast closed-loop reaction is necessary; action-chunk-based control is preferred for coordination-heavy or high precision tasks where smoothing mitigates jitter and limits error compounding (*13*). Both regimes share the same diffusion backbone; only the action heads differ.

Because we target deployment in homes and factories, we emphasize deployment-centric metrics: **reliability** (success rate), **efficiency** (time-to-completion), and **robustness** (sustained stability under long deployment time and perturbation). Real-world experiments show that RL-100 attains 100% reliability across all eight tasks and maintains long-horizon stability. In terms of efficiency, RL-100 approaches human teleoperation-level time-to-completion and, on several tasks, *matches or even surpasses* skilled operators. In summary, our main contributions are as follows:

(i) **Unified training framework.** We propose RL-100, a three-stage real-world RL training framework on top of teleoperation-trained generative diffusion policies. The pipeline chains IL pretraining, iterative offline RL and online RL, with most updates allocated to iterative offline and a small, targeted online budget for the last mile to deployment-grade performance.

(ii) **One objective, fast deployment.** A unified policy gradient loss fine-tunes the diffusion sampler's denoising schedule; a lightweight distillation compresses the multi-step diffusion policy to a one-step consistency policy.

(iii) **Generality across tasks, embodiments, and visual representations.** Our framework is task-, embodiment-, and visual-representation–agnostic. To the best of our knowledge, RL-100 is the first system to demonstrate vision-based RL post-training on real robots across diverse task modalities and embodiments.

(iv) **Deployment-centric results.** On real robots, RL-100 achieves 100% success across eight tasks (including runs of up to 250+ consecutive trials) and matches or exceeds human teleoperation efficiency on multiple tasks. Notably, the juicing robot served random customers continuously for about seven hours without failure under zero-shot shopping mall deployment.

(v) **RL-specific network backbone.** Our policy backbone is tailored for diffusion-based visuomotor control and is agnostic to the execution regime, supporting both single-step and action-chunk control. We use a self-supervised visual encoder to produce stable, drift-resistant representations throughout RL fine-tuning.

The whole system leverages RL to train diffusion models to solve robotic manipulation. Therefore, we survey related works on RL methods, generative diffusion models as well as other methods that address real-world robotic manipulation tasks. See supplementary for more details.

## Generative Diffusion Models in Robotics

The application of diffusion models to robotics has yielded transformative advances in visuomotor control, trajectory planning, and real-world deployment. Diffusion Policy (*3*) demonstrated breakthrough performance in visuomotor control by modeling action distributions as conditional diffusion processes, excelling at handling multimodal behaviors through visual conditioning and receding horizon control. This approach has inspired extensions including DiffClone (*31*) and Diffusion Model-Augmented Behavioral Cloning (*32*), which further improve upon traditional behavioral cloning by leveraging the expressiveness of diffusion models. The integration with pretrained visual representations, including R3M (*33*), VC-1 (*34*), and MoCo (*35*), has proven particularly effective for generalization across visual variations. Recent work on 3D Diffusion Policy (*4*) extends these capabilities to point cloud observations, while FlowPolicy (*36*) introduces consistency flow matching for robust 3D manipulation tasks. (*37*) propose a triply-hierarchical diffusion policy that decomposes complex visuomotor tasks into multiple levels of abstraction, improving both learning efficiency and generalization.

Beyond direct policy learning, diffusion models have revolutionized trajectory planning in robotics. (*38*) and (*39*) reformulate planning as conditional generation, producing complete state-action trajectories conditioned on rewards and constraints, excelling in long-horizon tasks requiring complex coordination. (*40*) introduces an alternative paradigm by generating policy parameters rather than trajectories in latent spaces, offering computational advantages. Real-world deployment challenges have been addressed through methods like One-Step Diffusion Policy (*41*), which uses distillation to achieve real-time performance suitable for robotic control. Successful applications now span deformable object manipulation with PinchBot (*42*), multi-task learning, and navigation scenarios. The availability of large-scale datasets like BridgeData V2 (*43*) and foundation models such as $\pi_0$ (*5*) enables broader generalization across robotic platforms and tasks, accelerating the transition from laboratory demonstrations to practical robotic systems.

## Real-world RL

Real-world RL trains directly on real robot dynamics, optimizing deployment metrics (reliability, speed, safety) and yielding robust performance that continually adapts to disturbances—without sim-to-real gaps. Critical requirements include *sample efficiency*, stability under high-dimensional perception, safe continuous operation, and automated reward/reset mechanisms. While early work demonstrated end-to-end visuomotor learning (*53*) and large-scale grasping with off-policy methods (*54*), subsequent advances established key algorithmic foundations: off-policy actor-critic methods (SAC, TD3) for data efficiency (*55*, *56*), model-based approaches for sample acceleration (*57*, *58*), reset-free learning for autonomous operation (*59*, *60*), and learned reward specifications from visual classifiers or human feedback (*61*, *62*). Despite these advances, most systems required extensive engineering, task-specific tuning, or long training times to achieve reliable performance.

These scattered advances converge in SERL (*9*), a comprehensive framework that integrates high update-to-data ratio off-policy learning, automated resets, and visual reward specification to achieve several manipulation tasks. However, SERL relies solely on demonstrations and struggles with tasks requiring precision or recovery from failures. HIL-SERL (*2*) addresses these limitations by incorporating real-time human corrections during training, enabling the policy to learn from mistakes and achieve perfect success rates across diverse tasks, including dual-arm coordination and dynamic manipulation.

While SERL and HIL-SERL report impressive on-robot learning efficiency and reliability on well-scoped tabletop skills, their evaluations typically employ action−space shaping (e.g., limiting wrist rotations and encouraging near-planar end-effector motion) and focus on short-horizon regimes with relatively low-dimensional control (*2*, *9*). Such constraints are pragmatic for safety and sample efficiency, but they reduce policy expressivity and can cap performance on orientation-critical, contact-rich, or compositionally complex tasks. In everyday home and factory scenarios, many skills inherently require full SE(3) control and substantial reorientation, including: deformable manipulation with twist and regrasp (e.g., towel folding and box folding, where small pose errors can compound into irreversible misfolds), insertion/ejection in confined cavities with large tilt changes (e.g., orange juicing), fluid and granular control that hinges on container tilt (e.g., controlled pouring), dynamic release and trajectory shaping (e.g., agile bowling), cable routing or cloth placement with out-of-plane rotations, and bimanual reorientation. *We do not claim these systems cannot be extended to such settings; rather, the reported experiments emphasize rapid learning on well-scoped tasks, leaving broader generalization, long-horizon composition, and orientation-sensitive control less explored.*

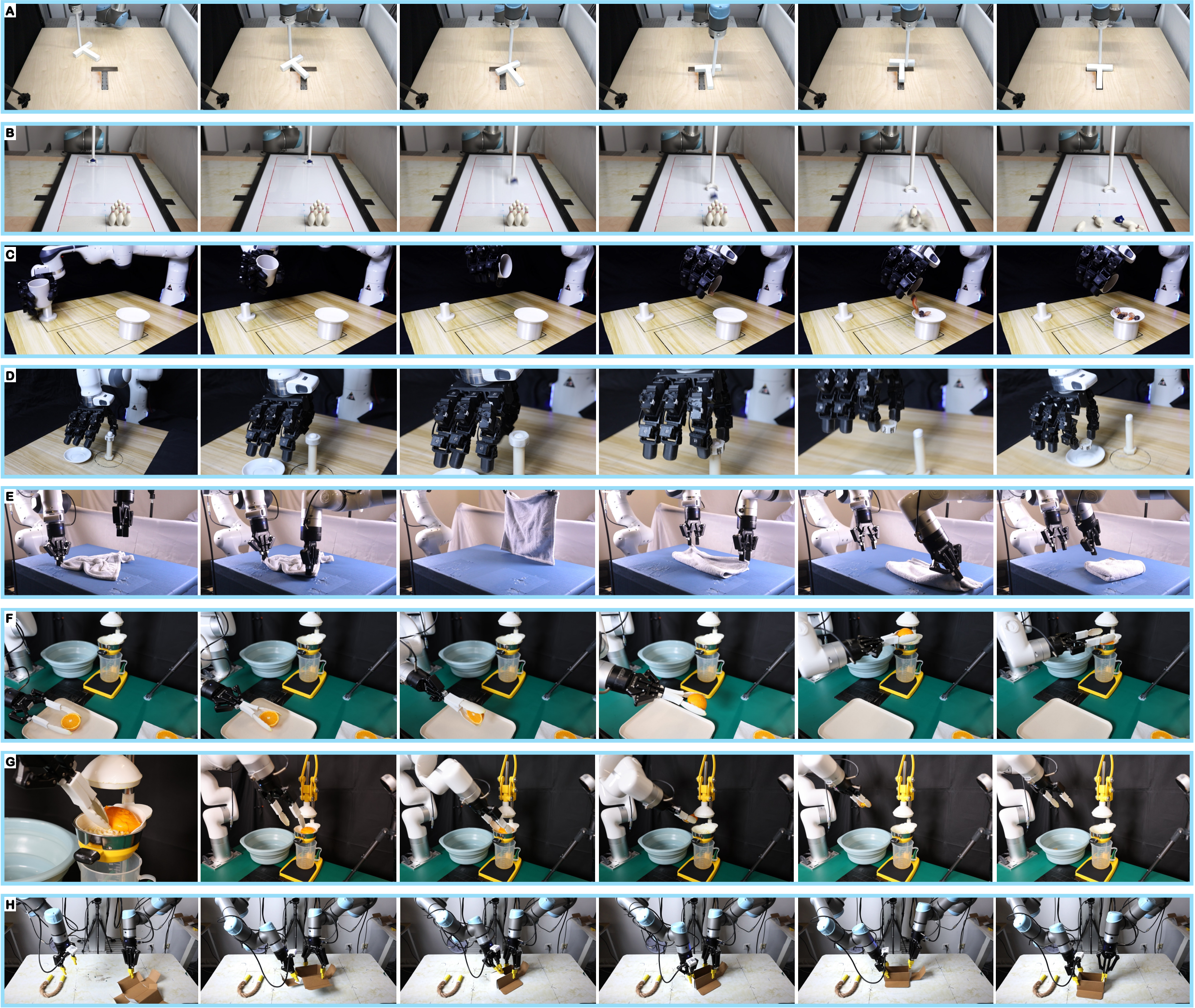

**Fig. 2. Rollout trajectories of eight real-world tasks.** Each row shows one task, and columns are time-ordered frames (left → right) subsampled from a single trajectory. From top to bottom: (A) Dynamic Push-T, (B) Agile Bowling, (C) Pouring, (D) Dynamic Unscrewing, (E) Soft-towel Folding, (F) Orange Juicing - Placing, (G) Orange Juicing – Removal, (H) Box Folding. The suite spans fluids or particles, tool-using, deformable objects manipulation, dynamic non-prehensile manipulation, and precise insertion, highlighting the diversity and dynamics of our benchmark.

In contrast, our system retains full 6-DoF control without hard rotation constraints and targets these under-explored regimes by (i) using a diffusion/consistency visuomotor policy to capture diverse human strategies, (ii) unifying offline-to-online improvement with an offline policy evaluation (OPE)-gated clipped PPO surrogate objective for nearly-monotonic improvement, and

(iii) enabling high-frequency control via one-step consistency distillation, across dual-arm, deformable, and dynamic tasks with larger cross-object generalization.

# REAL-WORLD RESULTS

In this section, we detail the experimental setup and report results on reliability (success rate), efficiency (time-to-completion), and robustness (generalization across objects and initial conditions). We compare policies trained with RL-100 to human teleoperators and strong baselines. Across tasks, RL-100 achieves **higher precision and greater efficiency than human teleoperation**. We then present ablations to analyze the contribution of each component of RL-100. Collectively, these results demonstrate the **practical viability of RL-100 for real-world deployment.** See the Supplementary Materials (Simulation Results and Analysis) for additional analyses.

## Overview

We evaluate RL-100 on eight real-world manipulation tasks that jointly cover dynamic rigid-body control, deformable-object handling, and precision assembly, as shown in Fig. 1 and Table S1, to comprehensively evaluate the versatility of our framework. The suite spans single- and dual-arm embodiments (UR5, Franka with LeapHand, xArm–Franka) and two control modes (single-step actions vs. chunked action sequences). Besides randomized initial layouts shown in Fig. S2 to encourage spatial generalization, each task features unique challenges ranging from dynamic non-prehensile manipulation, stochastic interaction, deformable object, appearance variation, etc. Therefore, we do not maximize initial randomization but emphasize the combination between spatial layouts and task-specific difficulties. The trajectory of each task visualized in images and point clouds can be found in Fig. 2 and Fig. S1, respectively.

Key specifications for observation and action spaces of all tasks are summarized in Table S2. We randomize initial object placements for each trial to encourage policy generalization, with specific ranges for each task depicted in Fig. S2. Two types of reward structure are used for our tasks. For all tasks except Dynamic Push-T, we use ***a sparse reward function*** where the agent receives a reward of +1 upon successful task completion and 0 at all other timesteps, labeled by a human supervisor pressing a keyboard. To further optimize execution efficiency (i.e., reduce time-to-completion) and stabilize control, we add (i) a small per-timestep step penalty that discourages unnecessarily long trajectories, and (ii) a penalty on action jitter, implemented as the deviation between the current action and a moving average of recent actions over a fixed temporal window, which suppresses high-frequency fluctuations in the control signal. The Dynamic Push-T task, which requires continuous, high-precision control, utilizes a dense shaped reward. The total reward at each timestep $t$ is $r_t = r_{\text{pose}} + r_{\text{static}} + r_{\text{smooth}}$, where $r_{\text{pose}} = \exp(-3e) - 1$ punishes the SE(3) discrepancy $e$ between the T-block and the desired pose; $r_{\text{static}} = -1$ if the movement of T-block at one timestep is below a given threshold and otherwise 0; $r_{\text{smooth}} = -5\|a_t - a_{t-1}\|_2^2$ punishes jerky actions.

Across tasks, RL-100 achieves higher success rates and shorter time-to-completion than baselines and human operators, with especially large gains in settings that require fast online corrections (dynamic contacts, narrow clearances) or stable dexterous grasping under pose variability. A comprehensive robustness protocol evaluates either zero-shot or few-shot transfer across challenging variations (e.g., changed objects, visual distractors, external perturbations). Results show consistent performance retention under these shifts. We defer task-specific objectives and success criteria to the following section, and present aggregate reliability/robustness/efficiency results afterwards.

**Task descriptions and evaluation protocols.** Due to space constraints, we defer the full per-task descriptions to the Supplementary Materials (Description of Tasks). Specifically, it provides the objective, success criteria, key challenges, initialization distributions, and additional evaluation protocols/variants for each task (Dynamic Push-T, Agile Bowling, Pouring, Dynamic Unscrewing, Soft-towel Folding, Orange Juicing, and Box Folding). We also detail task-specific robustness and generalization tests (e.g., surface/friction changes, visual distractors, external disturbances, zero-shot transfers, and out of distribution (OOD) initialization regimes).

## Main Results

Table 1 traces performance from imitation-only policies to our RL-100 method. The imitation baselines yield only moderate results overall (mean success: 45.3% for DP-2D; 67.8% for DP3). DP-2D struggles on dynamic and contact-rich tasks (e.g., Agile Bowling at 14% and Pouring at 42%), while DP3 improves substantially on some tasks (e.g., Agile Bowling to 80%) but remains challenged on long-horizon/contact-rich manipulation such as Pouring (48%) and Box Folding (48%). In particular, both imitation baselines underperform on Box Folding (12% for DP-2D; 48% for DP3), consistent with the task's long-horizon bimanual coordination, error-sensitive crease/edge alignment, and complex, discontinuous contact transitions when folding a semi-rigid deformable object.

The iterative offline RL phase of RL-100 yields a large improvement and raises mean success to 91.8%. Relative to DP3, it delivers the largest absolute gains on the hardest tasks—most notably Box Folding by +48 points (96% vs 48%) and Pouring by +44 points (92% vs 48%)—while also improving Dynamic Unscrewing by +24 points (94% vs 70%), and consistently boosting other tasks (e.g., +26 points on Dynamic Push-T and Soft-towel Folding). These results suggest that conservative, model-guided policy refinement is particularly effective for long-horizon, contact-rich tasks where errors compound, while avoiding the instability often associated with naive online updates.

**Table 1. Success rate (%) across tasks.** RL-100 (ours) groups an iterative offline fine-tuning stage followed by online RL with either DDIM or one-step CM policy. Averages are unweighted across tasks.

| Task | Imitation baselines | | RL-100 (ours) | | |
|---|---|---|---|---|---|
| | DP-2D | DP3 | Iterative Offline RL | Online RL (DDIM) | Online RL (CM) |
| **Dynamic Push-T** | 40 (20/50) | 64 (32/50) | 90 (45/50) | 100 (50/50) | 100 (50/50) |
| **Agile Bowling** | 14 (7/50) | 80 (40/50) | 88 (44/50) | 100 (50/50) | 100 (50/50) |
| **Pouring** | 42 (21/50) | 48 (24/50) | 92 (46/50) | 100 (50/50) | 100 (50/50) |
| **Soft-towel Folding** | 46 (23/50) | 68 (34/50) | 94 (47/50) | 100 (50/50) | 100 (250/250) |
| **Dynamic Unscrewing** | 82 (41/50) | 70 (35/50) | 94 (47/50) | 100 (50/50) | 100 (50/50) |
| **Orange Juicing — Placing** | 78 (39/50) | 88 (44/50) | 94 (47/50) | 100 (100/100) | 100 (50/50) |
| **Orange Juicing — Removal** | 48 (24/50) | 76 (38/50) | 86 (43/50) | 100 (50/50) | –* |
| **Box Folding** | 12 (6/50) | 48 (24/50) | 96 (48/50) | 100 (50/50) | 100 (50/50) |
| **Mean (unweighted)** | **45.3** | **67.8** | **91.8** | **100.0** | **100.0†** |

†Mean over the seven tasks evaluated with CM. *Juicing - Removal exhibits IK-induced pose discontinuities between push and grasp in a tight, slippery contact; this issue is already present in the demonstrations and is exacerbated by the noise-sensitive one-step CM policy; for safety, we did not evaluate CM on this task.

The full RL-100 pipeline, consisting of iterative offline RL followed by brief on-policy online refinement, attains perfect success across all evaluated trials. With the DDIM policy, we observe 450 successful episodes out of 450 across eight tasks (including the two Orange Juicing subtasks). The consistency-model variant achieves the same perfect success rate on seven evaluated tasks (550/550) and enables single-step high-frequency control, including 250 successful trials out of 250 on the challenging dual-arm Soft-towel Folding task. Overall, the results highlight the contribution of each component: OPE-gated iterative learning delivers substantial and reliable improvements under a conservative regime, and short on-policy fine-tuning together with consistency distillation removes remaining failure modes and produces computationally efficient deployment-ready policies.

## Zero-shot Adaptation to New Dynamics

To assess the robustness of our learned policies, we evaluate zero-shot transfer to environmental variations unseen during training, listed in Table 2 (top part) and Fig. 3A-F. On Dynamic Push-T, as shown in Fig. 3B-C, the policy maintains perfect 100% success when the surface friction is substantially altered and achieves 80% success even when visual and physical clutter from interference objects of various shapes are introduced into the workspace. Similarly, Agile Bowling in Fig. 3D achieves 100% success on modified surface properties that change ball rolling dynamics. For Pouring in Fig. 3A, replacing granular nuts with fluid water—a significant shift in material properties and flow dynamics—still yields 90% success, demonstrating that the policy has learned sufficiently stable and robust manipulation strategies rather than overfitting to specific physical parameters. For soft-towel folding on unseen shape in Fig. 3E, RL-100 attains 80% success, suggesting insensitivity to towel geometry and robust representation learning. For Box Folding in Fig. 3F, we evaluate an extreme OOD initialization regime with unseen blank shapes and near-orthogonal orientations to the bimanual manipulation axis (large $\theta$), which are rarely present in the training demonstrations. RL-100 achieves 90% success, indicating robust long-horizon bimanual reorientation and crease-accurate folding under severe geometric distribution shift. Across all four variations, our method achieves 90% average success without any retraining or fine-tuning, indicating strong generalization to distribution shifts in environmental dynamics.

## Few-shot Adaptation

Beyond zero-shot generalization, we evaluate the sample efficiency of adapting to more substantial task modifications through brief fine-tuning, as shown in Table 2 (middle part) and Fig. 3G-I. After only 1-3 hours of additional training on each variation, our policies achieve high success rates across diverse changes. The Soft-towel Folding policy adapts perfectly to a different towel material with altered deformable properties, achieving 100% success rate as shown in Fig. 3G, demonstrating effective transfer despite significant changes in cloth dynamics. Similarly, Agile Bowling achieves 100% success on an inverted pin arrangement listed in Fig. 3I, requiring the policy to adapt its trajectory planning and aiming strategy. The Pouring task with a modified container geometry, as shown in Fig. 3H, achieves 60% success, showing that while geometric changes require more adaptation, the policy can still leverage its learned manipulation primitives. Averaging 86.7% success across these variations with minimal retraining demonstrates the sample efficiency and adaptability of our approach, highlighting that the learned representations and control strategies effectively transfer to modified task configurations.

## Robustness to Physical Disturbances

We evaluate the resilience of RL-100 policies to real-world physical perturbations by introducing human-applied disturbances during task execution, as shown in Table 2 (bottom part) and Fig. 3J-Q. For Soft-towel Folding, we apply external forces at two critical stages: during initial grasping (Stage 1, Fig. 3J) and during pre-folding manipulation (Stage 2, Fig. 3K), achieving 90% success at both stages despite the perturbations. On the Dynamic Unscrewing task (Fig. 3L), we apply sustained counter-rotational interference for up to 4 seconds during both the unscrewing motion and at the critical zero-boundary point where visual judgment is required to grasp the nut, a particularly challenging moment as described in our task specifications. Remarkably, the policy achieves 100% success, demonstrating the ability to recover from prolonged disturbances even during precision-critical manipulation phases. Similarly, Dynamic Push-T (Fig. 3M) achieves 100% success despite multiple dragging perturbations applied throughout the pushing motion. For Box Folding (Fig. 3N–Q), RL-100 maintains 100% success under a combined disturbance suite spanning two grasp-stage perturbations, push interference, and repeated touch/tap contacts.

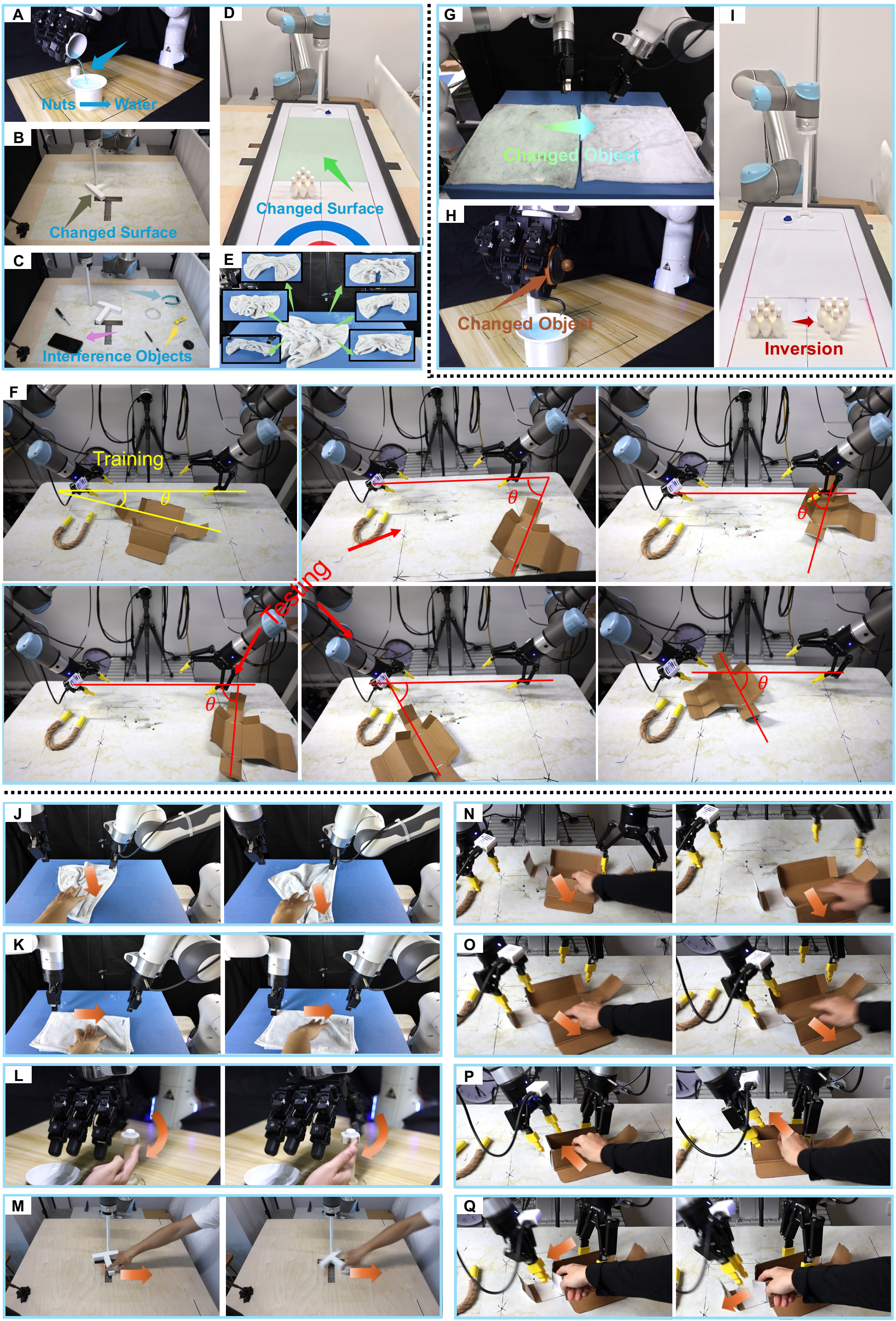

**Fig. 3. Evaluating our policies under various task variations.** (A-F) Zero-shot adaptation to novel dynamics without retraining. (G-I) Few-shot adaptation to unseen scenarios with minimal additional training data. (J-Q) Robustness against external disturbances. (A) Pouring with granular/fluid materials. (B) Dynamic Push-T on altered surface friction. (C) Dynamic Push-T with visual and physical interference objects. (D) Agile Bowling on modified surface. (E) Soft towel Folding generalizes to unseen shapes; green arrows mark the diverse test-time configurations. (F) Box Folding with unseen shape and orientation (G) Soft-towel Folding with a different towel exhibiting altered deformable properties. (H) Pouring with a modified container shape and size. (I) Agile Bowling with inverted pin arrangement. (J-K) Soft-towel Folding recovers from external pulling and lateral dragging at different manipulation stages. (L) Dynamic Unscrewing withstands counter-rotational interference from human hand. (M) Dynamic Push-T handles dragging perturbations to the block. Arrows indicate variations or disturbances. (N-Q) Box Folding with human disturbances during execution, including dragging, pushing, and tapping.

Across all tested scenarios, our policies maintain an average 96.0% success rate, validating that the closed-loop control learned through RL-100 enables robust recovery and task completion in the presence of external perturbations typical of unstructured real-world environments.

**Table 2. Success rates (%) on different task variations with novel dynamics and environmental variations with zero-shot adaptation (top part), few-shot adaptation (middle part) and physical disturbances (bottom part).** In zero-shot and disturbed cases, all policies are trained on nominal conditions and evaluated without any retraining or fine-tuning. In few-shot cases, policies are finetuning 1-3 hours.

| Type | Task variation | Success Rate (%) |
|---|---|---|
| Zero-shot Adaptation | Pouring (water) | 90 |
| | Push-T (changed surface) | 100 |
| | Push-T (interference objects) | 80 |
| | Bowling (changed surface) | 100 |
| | Soft-towel Folding (Unseen shape) | 80 |
| | Box Folding (Unseen shape/Orientation) | 90 |
| | **Average** | **90.0** |
| Few-shot Adaptation | Pour (new container) | 60 |
| | Folding (changed object) | 100 |
| | Bowling (inverted pin) | 100 |
| | **Average** | **86.7** |
| Robustness against disturbances | Soft-towel Folding @ Grasping | 90 |
| | Soft-towel Folding @ Pre-folding | 90 |
| | Unscrewing | 100 |
| | Push-T | 100 |
| | Box Folding | 100 |
| | **Average** | **96.0** |

## Execution Efficiency

We evaluate deployment-time execution efficiency, i.e., how quickly a deployed policy completes tasks on real hardware, using four complementary metrics (Fig. 4): (A) environment steps per successful episode, (B) end-to-end wall-clock time per successful episode, (C) sustained throughput under consecutive execution on Dynamic Push-T, and (D) average episode length including failures/timeouts on Dynamic Push-T. We report (A–B) on action-chunk tasks where policy-side inference can dominate latency, and (C–D) on Dynamic Push-T where repeated trials and in-episode recovery make throughput and failure-aware efficiency most informative.

**Episode length in successful trajectories.** We first compare the number of environment steps needed to complete a task successfully (Fig. 4A). RL fine-tuned policies (RL-100) consistently outperform imitation baselines across all three tasks. This advantage stems from reward-driven policy optimization, which steers the agent toward more efficient—and potentially optimal—trajectories for earlier task completion. Specifically, RL-100 achieves substantial step reductions: in *Soft-towel Folding*, 390 steps (DP-2D) → 312 steps (RL-100 CM; 1.25× fewer), and in *Dynamics Unscrewing*, 361 steps (DP-2D) → 280 steps (RL-100 DDIM; 1.29× fewer). The largest gain appears in Box Folding, a substantially long-horizon, contact-rich task: 1266 steps (DP-2D) → 832 steps (RL-100 CM; 1.52× fewer). These gains reflect the value-driven objective's tendency to produce more decisive state transitions.

**Wall-clock time per episode.** Beyond step count, wall-clock time is the most intuitive metric for real-world deployment, especially for long-horizon real-robot tasks where sensing and policy inference contribute non-trivial latency. Fig. 4B reports **end-to-end wall-clock time**, i.e., the effective **deployment-time latency**, which depends on two factors: (i) *Input modality*—point clouds (DP3, 1024 points) incur lower computational overhead than 128 × 128 RGB images (DP-2D), yielding 1.02−1.31× speedup despite similar step counts; (ii) *Inference method*—RL-100 (CM)'s one-step generation per action reduces control-loop latency compared to the 10-step DDIM sampler, providing an additional 1.05−1.16× wall-clock speedup over RL-100 (DDIM). This effect is most pronounced on Box Folding, where imitation baselines can exceed one minute: DP-2D requires 65.1s, while RL-100 (CM) completes in 41.4s (1.57× faster) and RL-100 (DDIM) in 45.6s, demonstrating that RL fine-tuning plus one-step distillation yields tangible deployment-time gains on long-horizon tasks.

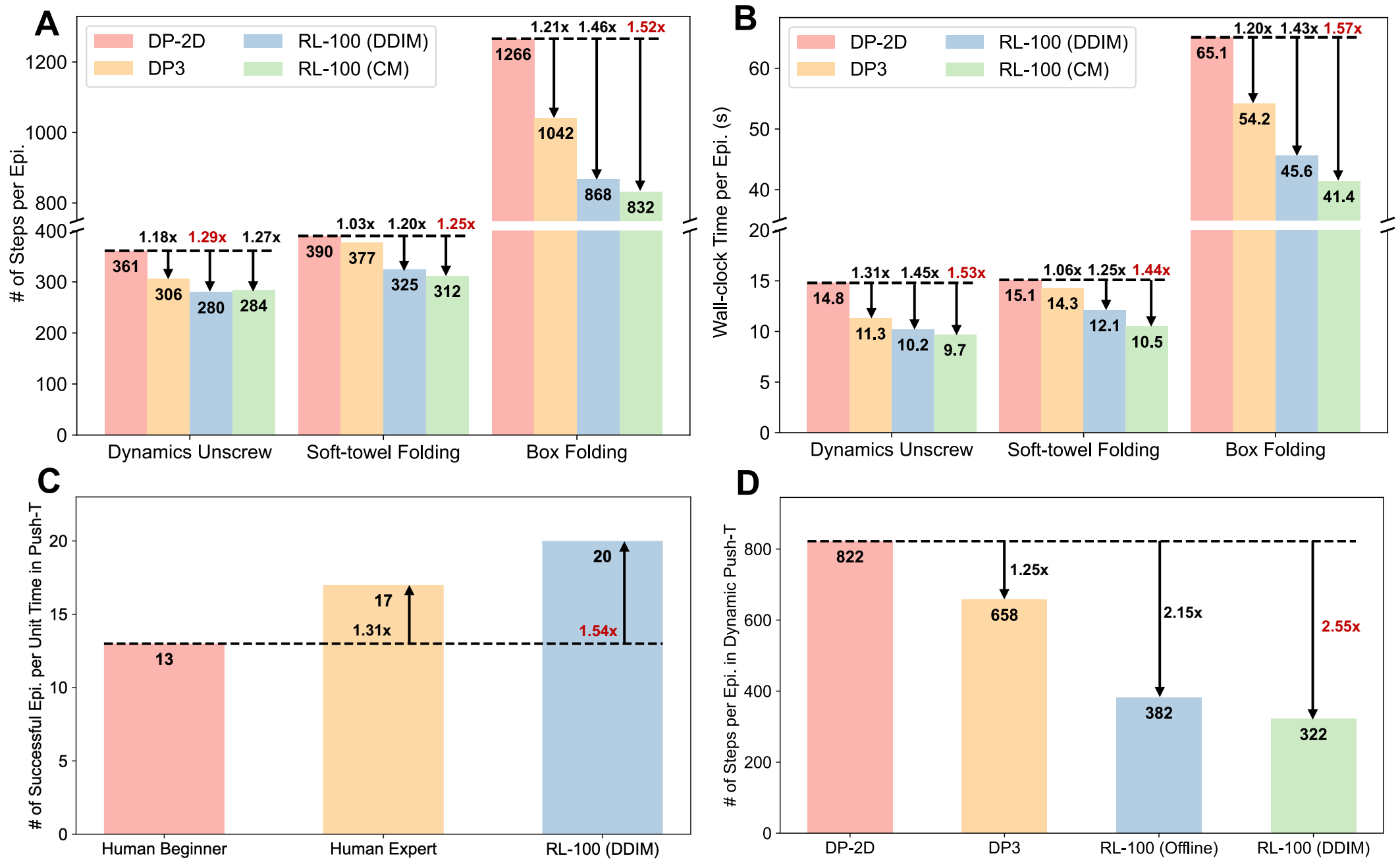

**Fig. 4. Comparison of RL-100 against algorithmic baselines or human operators in terms of deployment-time task completion efficiency.** (A) Episode length (environment steps) conditioned on success, averaged over 50 successful episodes per method for each task (lower is better). (B) End-to-end wall-clock time (s) per successful episode, averaged over the same 50 successful episodes (lower is better). Panels (A–B) report results on Dynamic Unscrewing, Soft-towel Folding, and Box Folding. (C) Dynamic Push-T throughput measured under consecutive execution: wall-clock time to reach 20 successes, reported as 20/time (successes per unit time; higher is better). (D) Episode length in Dynamic Push-T including failures/timeouts, averaged over 50 episodes with a 1000-step episode limit (lower is better). Bars show absolute values; annotations indicate speedup factors relative to the baseline (DP-2D, DP3 for A, B, and D; Human Beginner/Expert for C). We abbreviate ‘episode’ as ‘epi.’.

**RL-100 vs. human operators.** A critical question for practical deployment is whether robotic policies can match or exceed human efficiency. To address this, we compare RL-100 against human teleoperators on Dynamic Push-T (Fig. 4C). RL-100 (DDIM) achieves 20 successful episodes per unit time, surpassing the human expert who collected the demonstrations (17 episodes; 1.18× improvement) and substantially exceeding human beginners (13 episodes; 1.54× improvement). This demonstrates that our learned policy not only matches expert-level task completion quality but also executes more efficiently in repetitive scenarios.

**Episode length including failures.** The previous comparisons focused solely on successful trajectories. However, a comprehensive efficiency analysis must account for all attempts, including failures. Fig. 4D shows that when considering all episodes including failures/timeouts on Dynamic Push-T, RL-100 maintains significantly shorter average horizons: 822 steps (DP-2D) → 658 steps (DP3) → 382 steps (RL-100 Offline) → 322 steps (RL-100 DDIM), representing up to 2.55× fewer steps than DP-2D. This indicates that RL-100 produces policies that both succeed more reliably and terminate unpromising attempts quickly, avoiding wasted execution time.

**Summary.** Overall, the task completion efficiency gains of RL-100 arise from three synergistic sources: (i) *efficient encoding* (point clouds over RGB images), (ii) *reward-driven policy optimization* that encourages faster completion and reduces dithering, and (iii) *low-latency action generation* (CM's one-step versus DDIM's 10-step inference). The monotonic improvement—DP-2D → DP3 → RL-100 (DDIM) → RL-100 (CM)—validates each design choice's contribution. **Crucially, RL-100's execution efficiency surpasses both algorithmic baselines and human operators, demonstrating readiness for practical deployment.**

## Training Efficiency

Fig. S5 shows the online RL training progression for Agile Bowling. The policy achieves consistent 100% success after approximately 190 episodes of on-policy rollouts, with a moving average demonstrating stable learning without performance drop from offline. The final 100+ episodes maintain perfect success, validating robust policy convergence. *These success rates reflect training rollout episodes with exploration noise, not formal evaluations. Since rollouts involve noise for exploration, while evaluation uses deterministic actions, the reported rates underestimate the true policy performance. However, frequent evaluation on physical hardware is prohibitively costly. We therefore track learning progress through rollout success rates as an efficient proxy, providing real-time visibility into policy improvement. We present this curve for Agile Bowling as it exemplifies the rapid convergence and stability characteristic of our online RL fine-tuning phase.*

As shown in Table S3, our pipeline reaches deployment-grade performance with minimal human teleoperation and moderate policy rollouts. Across eight tasks, we average only 115 human demos (about 1.8 (h)) per task, after which iterative offline RL adds ~566 rollouts (6.5 (h)) and online RL adds about 434 rollouts (5.6 (h)) per task. Summed over all tasks, human collection is just 804 episodes in 12.5 (h)—about 13% of a total data-collection budget under 100 (h)—while the remaining time comes from autonomous policy rollouts (offline: 3965 eps/45.3 (h); online: 3037 eps/39.5 (h)). This balance demonstrates that a small amount of expert data, complemented by moderate, inexpensive rollouts, suffices to train policies that meet our deployment criteria.

## Comparison to Humans and Human–Robot Collaboration

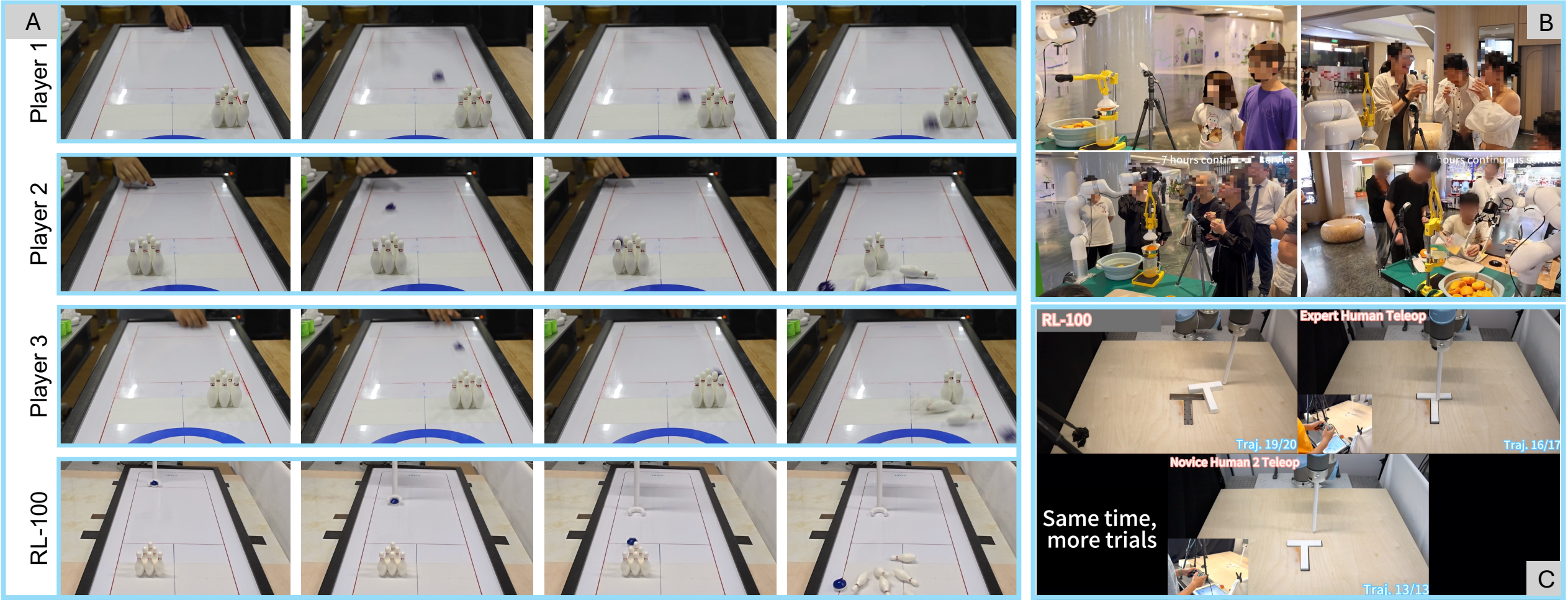

**Fig. 5. RL-100 in competitive and service settings.** (A) Bowling matches between RL-100 and several human players. (B) Public deployment in a shopping mall where RL-100 prepares fresh juice for visitors during a multi-hour demo. (C) Head-to-head comparison on a T-shaped block-pushing benchmark; RL-100 achieves more trials within the same time window.

Fig. 5 positions RL-100 relative to humans in both competitive and assistive settings, consistent with our deployment criteria (reliability, efficiency, robustness). In tabletop bowling (Fig. 5A), with 25 attempts per side, RL-100 achieves 25 successes versus 14 for human players. In a public shopping-mall demonstration (Fig. 5B), RL-100 prepares fresh juice continuously for seven hours without failure, evidencing robustness outside the lab and zero-shot transfer to a new environment. On the standardized T-push benchmark (Fig. 5C), under an equal wall-clock budget the robot completes 20 successful trials compared with 17 by an expert teleoperator and 13 by a novice, indicating lower time-to-completion. Together, these results illustrate that starting from strong human priors and post-training with real-world RL enables RL-100 to match−and in several tasks exceed−skilled human performance while remaining suitable for real deployments.

# DISCUSSION

The results advance the state of the art in robotic manipulation in terms of reliability, efficiency, and robustness. Across eight real-world tasks—dynamic pushing, dual-arm soft-towel folding, agile bowling, pouring (granular/fluids), unscrewing, and multi-stage juicing—it achieves 100% success in 1000/1000 trials. The policy generalizes across initial object placements and tolerates shape/size variations, while matching or surpassing expert teleoperators in time-to-completion. It remains stable in extended deployment, including a public mall demo running more than seven hours without failure, and sustains high performance under significant environmental shifts without retraining. Together, these results indicate that RL-100 meets deployment-level requirements.

In summary, RL-100 uses a small set of human demonstrations as a foundation and then aligns with human-grounded objectives, exceeding human-level reliability through iterative self-improvement. This reduces the burden on data-collection staff and further benefits from large volumes of autonomous on-robot rollouts. Its perfect success rates, broad task generality, and sustained operational robustness represent a significant step toward deploying autonomous manipulators in unstructured environments, outlining a credible path to deployment in homes and factories.

The findings suggest that small diffusion policies can attain high reliability and efficiency after modest fine-tuning. Because RL-100's training pipeline is model-agnostic, we can scale post-training to larger, multi-task, multi-robot diffusion/flow-based vision-language-action (VLA) models. Key directions include: (i) establishing scaling laws for data and model size versus real-robot sample efficiency; (ii) cross-embodiment and cross-task transfer within a single policy; and (iii) aligning large VLA priors with RL-100's unified objective to retain zero-shot semantic generalization while preserving the reported high success rates.

We will also extend evaluation to more complex, cluttered, and partially observable settings that better mirror homes and factories, including dynamic multi-object scenes, occlusions, specular/transparent materials, changing illumination, and non-stationary layouts. Although our pipeline already emphasizes conservative operation and stable fine-tuning, reset and recovery remain practical bottlenecks. We will explore autonomous reset mechanisms—learned reset policies, scripted recovery behaviors, task-aware fixtures, and failure-aware action chunking—to minimize on-robot human effort during training and evaluation.

# MATERIALS AND METHODS

## Methods

**Motivation.** Real-world manipulation tasks such as Box Folding and Orange Juicing involve contact-rich, long-horizon action sequences with sparse success signals, making learning from scratch on hardware prohibitively expensive and unsafe. We therefore start from *diffusion-based* ***imitation learning (IL)***, which provides a low-variance behavior prior grounded in human demonstrations and constrains exploration to a physically plausible action manifold (*3, 4*). *However,* *imitation alone does not directly optimize deployment objectives (success rate, time-to-completion, robustness to disturbances) and often fails to eliminate rare failure modes.*

***RL-based post-training*** offers a principled way to refine imitation policies into deployable controllers. Yet, naively applying online RL on top of IL often encounters distribution shift and overly aggressive updates, leading to performance crashes or sub-optimality (*47*). Recent behavior-centric policy optimization methods point to a remedy: BPPO (*48*) adopts the clipped importance-weighted surrogate from PPO to conservatively improve an imitation-initialized behavior policy in the ***offline regime***, and Uni-O4 (*47*) extends this idea by using the same clipped PPO surrogate objective to smoothly transition from offline refinement to on-policy ***online fine-tuning***. *Using a consistent proximal objective across regimes helps avoid performance dips and improves stability in practice, motivating a unified IL-to-RL post-training recipe.*

Extending this idea to modern ***visuomotor manipulation*** is nontrivial: diffusion-based policies generate actions through a multi-step denoising process with highly nonlinear, multimodal structure, and applying RL only at the final action yields weak credit assignment. *RL-100 addresses this by modeling denoising as a two-level MDP and sharing the environment-level advantage across denoising steps, enabling stable refinement of diffusion-based visuomotor policies in both offline and online regimes via the clipped PPO surrogate objective, while retaining their expressivity—crucial for high-dimensional, contact-rich tasks such as bimanual box folding.*

Overall, RL-100 is a three-stage pipeline, as illustrated in Fig. 6: (i) IL from demonstrations to initialize an expressive behavior prior, (ii) iterative offline RL with progressive data expansion to harvest most performance gains under conservative updates, and (iii) brief online fine-tuning to eliminate residual failures. The key innovation is a unified clipped PPO surrogate objective that applies consistently across diffusion denoising steps, enabling smooth offline-to-online refinement without changing the learning objective.

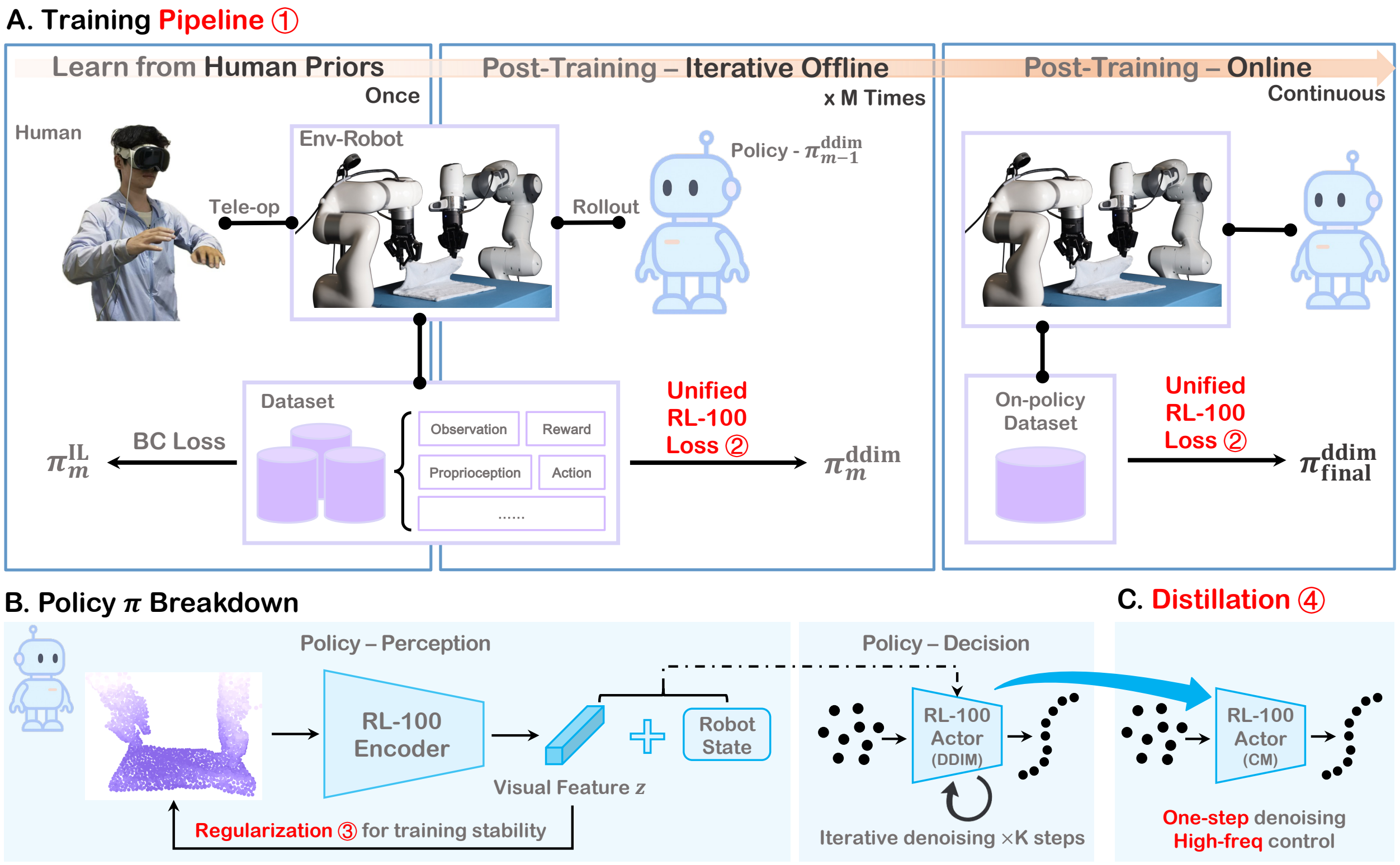

**Fig. 6. Overview of RL-100.** We mark key takeaways with numbers. (A) The framework features a three-stage training pipeline—imitation learning from human demonstrations, iterative offline post-training with data expansion, and continuous online post-training. A unified formulation

of RL-100 loss (extended from PPO) is used to update the policy in both offline and online stages. (B) A policy can be factorized into an observation encoder with a diffusion-based actor. We propose regularization techniques (reconstruction and variational information bottleneck) to stabilize post-training. (C) We design a consistency-model distillation to enable one-step action generation and high-frequency control at deployment.

### Imitation Learning

We initialize the policy by behavior cloning on human-teleoperated trajectories. Our approach uses conditional diffusion to learn robust visuomotor policies from demonstrations. Each episode provides synchronized tuples.

$$\{(o_t, q_t, a_t)\}_{t=1}^{T_e}, \tag{1}$$

where $o_t$ are visual observations (RGB images or 3D point clouds), $q_t$ denotes robot proprioception (joint positions/velocities, gripper state), and $a_t$ is either a single-step action or an action chunk.

**Conditioning and prediction horizon.** We fuse recent observations into a conditioning vector:

$$c_t = [\phi(o_i, q_i)]_{i=t-n_o+1}^{t}, \tag{2}$$

where the perception encoder $\phi(\cdot)$ processes the most recent $n_o$ frames (typically $n_o = 2$) and $[\cdot]$ is the operator concatenating multiple vectors. The clean diffusion target $a_t^{\tau_0}$ at timestep $t$ is set to a single action $a_t^{\tau_0} = u_t \in \mathbb{R}^{d_a}$ or an action chunk $a_t^{\tau_0} = [u_t, \dots, u_{t+n_c-1}] \in \mathbb{R}^{n_c d_a}$ where $n_c$ is the chunk size (typically 8-16). Actions are normalized per dimension; we predict delta end-effector pose when applicable.

**Diffusion parameterization.** Following conditional diffusion over actions, we corrupt $a_t^{\tau_0}$ to $a_t^{\tau_K}$ via the forward process (Equation (S2)) The denoiser $\varepsilon_\theta(a^\tau, \tau, c_t)$ is trained with the noise-prediction objective:

$$\mathcal{L}_{\text{IL}}(\theta) = \mathbb{E}_{(a^{\tau_0}, c_t)\sim\mathcal{D},\tau,\varepsilon}[\|\varepsilon - \varepsilon_\theta(a^\tau, \tau, c_t)\|_2^2], \tag{3}$$

where $\mathcal{D}$ is the demonstration dataset, $\tau \in \{\tau_K > \cdots > \tau_1\}$ indexes a $K$-step schedule, and $\varepsilon \sim \mathcal{N}(0, \mathbf{I})$. The policy backbone is shared across control modes; only the output head differs: $\mathbb{R}^{d_a}$ (single-step) or $\mathbb{R}^{n_c d_a}$ (chunked).

**Vision and proprioception encoders.** For RGB input, we use pretrained ResNet/ViT backbones; for point clouds input, we adapt the 3D encoder of DP3 (*4*) with reconstruction regularization for stability during RL fine-tuning. Visual embeddings are projected and concatenated with proprioceptive features to form $c_t$. All encoders are trained end-to-end with Equation (3).

When applicable, we add reconstruction (Recon) and variational information bottleneck (VIB) regularization:

$$\mathcal{L}_{\text{recon}} = \beta_{\text{recon}}(d_{\text{Chamfer}}(\hat{o}, o) + \|\hat{q} - q\|_2^2), \tag{4}$$

$$\mathcal{L}_{\text{KL}} = \beta_{\text{KL}}\text{KL}(\phi(z|o, s) \parallel \mathcal{N}(0, \mathbf{I})), \tag{5}$$

where $o$ and $q$ denote the observed point cloud and proprioceptive vector; $\hat{o}$ and $\hat{q}$ are reconstructed observations given encoded embedding $\phi(o, q)$; $d_{\text{Chamfer}}$ is Chamfer distance between two set of point clouds. The complete imitation learning objective becomes:

$$\mathcal{L}_{\text{total}}^{\text{IL}} = \mathcal{L}_{\text{IL}} + \mathcal{L}_{\text{recon}} + \mathcal{L}_{\text{KL}}. \tag{6}$$

During RL fine-tuning, we reduce $\beta_{\text{recon}}$ and $\beta_{\text{KL}}$ by a factor of 10 to allow policy improvement while maintaining representation stability.

**Inference and Control.** At deployment, $K$-step DDIM sampling (Equation (S5)) generates actions:

$$\hat{a}_t^{\tau_0} \leftarrow \mathrm{DDIM}_K(\varepsilon_\theta(\cdot,\cdot,c_t)).$$

Single-step control executes $u_t \leftarrow \hat{a}_t^{\tau_0}$ immediately; chunked control executes $[u_t, \dots, u_{t+n_c-1}] \leftarrow \hat{a}_t^{\tau_0}$ in the following $n_c$ timesteps. Single-step control excels in reactive tasks (e.g., dynamic bowling), while action chunking reduces jitter in precision tasks (e.g., assembly). Both modes share the same architecture, enabling task-adaptive deployment.

**Unified Offline and Online RL Fine-tuning**

While diffusion-based IL provides a strong visuomotor prior, it does not explicitly optimize deployment metrics or eliminate rare failures. To bridge this gap, we present a unified RL post-training framework designed to refine the IL-initialized policy. The primary technical hurdle in applying reinforcement learning to diffusion policies is the high-dimensional, multi-step nature of action generation, which typically results in sparse credit assignment if rewards are only applied to the final sampled action. We overcome this by reformulating the task as a hierarchical two-level MDP, which enables a dense learning signal across the entire denoising chain (see Supplementary Materials (Preliminaries: Diffusion Policy and RL Fine-tuning)). Crucially, by employing a unified clipped PPO surrogate formulation, we ensure a smooth functional transition from offline refinement to online exploration, preventing the performance collapses often associated with switching between distinct optimization regimes.

**Two-level MDP structure.** Our approach operates on two temporal scales: (i) **Environment MDP:** Standard robot control with state $s_t$, action $a_t$, reward $R_t$. (ii) **Denoising MDP:** $K$-step diffusion process generating each $a_t^{\tau_0}$ through iterative refinement. The denoising MDP is embedded within each environment timestep, creating a hierarchical structure where $K$ denoising steps produce one environment action. The formal two-level MDP formulation is provided in the Supplementary Materials (Preliminaries: Diffusion Policy and RL Fine-tuning). Moreover, our formulation assumes single-step execution for clarity; the extension to action chunking (including the chunk-level reward and discount) is provided in the Supplementary Materials (Preliminaries: Handling single action vs. action chunk).

**Unified PPO objective over denoising steps.** Given the two-level MDP structure, we optimize the PPO objective w.r.t. the $K$-step diffusion process at each timestep $t$ in iteration $i$ via the summation across all denoising steps $k$:

$$J_i(\pi) = \mathbb{E}_{s_t \sim \rho_{\pi_i}, a_t \sim \pi_i}\left[\sum_{k=1}^{K} \min(r_k(\pi)A_t, \mathrm{clip}(r_k(\pi), 1-\epsilon, 1+\epsilon)A_t)\right], \tag{7}$$

and the loss function is:

$$\mathcal{L}_{\mathrm{RL}} = -J_i(\pi), \tag{8}$$

where $r_k(\pi)$ is the per-denoising-step importance ratio and $A_t$ is the task advantage tied to the environment timestep $t$. Here, $\pi_i$ denotes the behavior policy at PPO iteration $i$, and $\rho_\pi$ is the (discounted) state distribution under current policy $\pi$. The key insight is to share the same environment-level advantage $A$ across all $K$ denoising steps, providing dense learning signals throughout the denoising process while maintaining consistency with the environment reward structure.

**Log-likelihood for PPO ratios.** During RL fine-tuning we use a *stochastic* DDIM transition so that each denoising step admits a well-defined Gaussian sub-policy and log-likelihood, enabling stepwise PPO ratios; deterministic sampling is used at deployment. We provide the DDIM sub-policy view and the corresponding log-likelihood in Equations (S5-S10).

**Offline and online stages (same objective, different data).** In the offline stage, we optimize Equation (8) using a fixed dataset $\mathcal{D}$ (initialized with demonstrations and progressively expanded with policy rollouts), computing the appropriate offline-policy ratios. In the online stage, we collect a small amount of on-policy data and continue optimizing the same objective to remove residual failures, benefiting from a smooth transition since the optimization primitive is unchanged across regimes.

**Algorithm 1** RL-100 training pipeline

**Input:** Demonstrations $\mathcal{D}_0$, iterations $M$
**Initialize:** $\pi_0^{\text{IL}} \leftarrow$ ImitationLearning($\mathcal{D}_0$)
**for** iteration $m = 0$ to $M - 1$ **do**
  // Offline RL improvement
  Train critics: $(Q_{\psi_m}, V_{\psi_m}) \leftarrow \text{IQL}(\mathcal{D}_m)$
  Train transition: $T_{\theta_m}(s'|s,a)$
  Optimize:
    $\pi_m^{\text{ddim}}, \pi_m^{\text{cm}} \leftarrow \text{OfflineRL}(\pi_m^{\text{IL}}, Q_{\psi_m}, V_{\psi_m}, T_{\theta_m})$
  // Data expansion
  Deploy: $\mathcal{D}_{\text{new}} \leftarrow \text{Rollout}(\pi_m^{\text{ddim}}$ or $\pi_m^{\text{cm}})$
  Merge: $\mathcal{D}_{m+1} \leftarrow \mathcal{D}_m \cup \mathcal{D}_{\text{new}}$
  // IL re-training on expanded data
  $\pi_{m+1}^{\text{IL}} \leftarrow$ ImitationLearning($\mathcal{D}_{m+1}$)
**end for**
// Final online fine-tuning
$\pi_{\text{ddim}}^{\text{final}}, \pi_{\text{cm}}^{\text{final}} \leftarrow \text{OnlineRL}(\pi_{M-1}, V_{\psi_{M-1}})$
**Output:** $\pi_{\text{ddim}}^{\text{final}}, \pi_{\text{cm}}^{\text{final}}$

## Iterative Offline RL

Rather than applying offline RL once on a fixed demonstration set, RL-100 uses an iterative procedure that alternates between conservative offline improvement and dataset expansion (Algorithm 1). Concretely, we start from demonstrations $\mathcal{D}_0$. At each outer iteration, we (i) train critics and a transition model on the current dataset, (ii) improve the current behavior policy via conservative offline RL updates, (iii) collect new rollouts with the improved policy, (iv) merge the new data into the dataset, and (v) re-train the diffusion policy with IL on the expanded dataset. This creates a virtuous cycle where better policies generate higher-quality data, which in turn enables learning even better policies.

Algorithm 1 uses an *outer* offline iteration index $m \in \{0, \ldots, M-1\}$ to denote dataset-expansion rounds. Within each outer iteration $m$, we run an *inner* sequence of PPO updates indexed by $i$, where $\pi_{m,i}$ denotes the current behavior policy used in the denominator of the PPO ratio; an offline policy evaluation (OPE) gate decides whether to accept an update and advance the behavior policy. We denote the behavior policy at the start of iteration $m$ as $\pi_{m,0} = \pi_m^{IL}$, and write $\pi_{m,i}$ as the behavior policy after the $i$-th accepted (OPE-gated) inner update. *For a fixed environment step $t$, we omit the subscript $t$ on denoising states/actions ($a_t^{\tau_{k-1}}, s_t^k$) for readability. See Supplementary Materials (Preliminaries) for the definition of diffusion schedule $\tau_k$.*

**Policy improvement on $\mathcal{D}_m$.** At offline inner iteration $i$, we optimize the clipped PPO surrogate objective (Equation (7)) applied across $K$ denoising steps using the ratio against the current behavior policy $\pi_{m,i}$:

$$r_k^{\text{off}}(\pi) = \frac{\pi(a^{\tau_{k-1}} \mid s^k)}{\pi_{m,i}(a^{\tau_{k-1}} \mid s^k)},$$

with standard clipping. $\pi$ is the candidate updated policy. Offline advantages are computed as:

$$A_t^{\text{off}} = Q_\psi(s_t, a_t) - V_\psi(s_t),$$

where the critics $(Q_\psi, V_\psi)$ are pre-trained on $\mathcal{D}_m$ following IQL (*68*) which is known for its conservative advantage estimation in offline settings. This yields a candidate policy $\pi$ from $\pi_{m,i}$ by several epochs of gradient updates on $\mathcal{D}_m$.

**OPE gate and iteration advancement.** In the inner-loop updates, naively iterating clipped-surrogate PPO updates can push the policy toward actions that are poorly supported by $\mathcal{D}_m$, where value/advantage estimates are unreliable (*47*). Such over-optimistic updates cause sharp performance crashes. We therefore use approximate model-Q function (AM-Q) (*47*) as an OPE gate, advancing the behavior policy $\pi_{m,i}$ only when the candidate is conservatively predicted to improve, yielding stable, OPE-monotonic behavior-policy improvement across inner iterations.

We use AM-Q for offline policy evaluation (OPE) without further interaction with the environment to compare update candidate policies with current behavior policy:

$$\hat{J}^{\text{AM-Q}}(\pi) = \mathbb{E}_{(s,a)\sim(\hat{T},\pi)}\left[\sum_{t=0}^{H-1} Q_\psi(s_t, a_t)\right],$$

where $\hat{T}$ is a learned transition model. We accept the candidate and advance the behavior-policy iteration only if:

$$\hat{J}^{\text{AM-Q}}(\pi) - \hat{J}^{\text{AM-Q}}(\pi_{m,i}) \geq \delta, \tag{9}$$

by setting the updated policy as behavior policy: $\pi_{m,i+1} \coloneqq \pi$. Otherwise, we reject and keep the behavior policy unchanged ($\pi_{m,i+1} \coloneqq \pi_{m,i}$). In practice, we set $\delta = 0.05 \cdot |\hat{J}^{\text{AM-Q}}(\pi_{m,i})|$ for adaptive thresholding. Given the estimated advantage, the loss function of offline RL $\mathcal{L}_{\text{RL}}^{\text{off}}$ equals to Equation (8).

**IL re-training matters.** Re-training with IL (Algorithm 1 line 13) on the expanded dataset is crucial for: (i) *distribution shift*: IL adapts to the evolving data distribution as higher-quality trajectories are added; (ii) *stability*: supervised IL acts as a regularizer, anchoring the policy to the high-density regions of the expanded dataset $\mathcal{D}_{m+1}$ and preventing the catastrophic forgetting of complex manipulation priors during pure RL exploration. (iii) *multimodality*: IL preserves the diffusion policy's ability to represent multiple solution modes; and (iv) *distillation*: IL compacts both human demonstrations and RL-improved behaviors into a single deployable policy.

**Shared and frozen encoders.** To ensure stable representation learning and efficient computation, all components in our offline RL pipeline share the same fixed visual encoder $\phi$ pre-trained during imitation learning. During offline RL, we keep $\pi^{\text{IL}}$ fixed and only update the task-specific heads of each module.

## Online RL

After iterative offline RL, we perform a brief on-policy refinement on real hardware to target residual rare failures. Starting from the offline-improved diffusion policy $\pi_{M-1}$ we collect an on-policy buffer $\mathcal{D}_{\text{on}}$ by rolling out $\pi_{M-1}$. Each environment control step $t$ produces one executed action $a_t^{\tau_0}$ through a $K$-step denoising chain (see Supplementary Materials (Preliminaries: Diffusion Policy and RL Fine-tuning)). *We omit the subscript* $t$ *on denoising states/actions* ($a_t^{\tau_{k-1}}, s_t^k$) *for readability.*

We compute an environment-level advantage $A_t^{\text{on}}$ using generalized advantage estimation (GAE) on finite-horizon trajectories with bootstrapping:

$$A_t^{\text{on}} = \text{GAE}(\lambda, \gamma; R_t, V_\psi),$$

where $V_\psi$ is the value function over environment states $s_t$ (warm-started from the offline critic and then updated on $\mathcal{D}_{\text{on}}$). Crucially, we share the same environment-level advantage $A_t^{\text{on}}$ across all $K$ denoising steps that generate $a^{\tau_0}$, providing dense learning signal throughout the denoising chain while remaining consistent with the environment reward.

We optimize the same clipped PPO surrogate objective with the policy ratio computed against a frozen behavior policy $\pi_i$:

$$r_k^{\text{on}}(\pi) = \frac{\pi(a^{\tau_{k-1}} \mid s^k)}{\pi_i(a^{\tau_{k-1}} \mid s^k)},$$

Then we minimize the total loss:

$$\mathcal{L}_{\text{RL}}^{\text{on}} = -J_i(\pi) + \lambda_V \mathbb{E}[(V_\psi(s_t) - \hat{V}_t)^2], \tag{10}$$

Where $J_i(\pi)$ is given by Equation (7) with the on-policy ratio $r_k^{\text{on}}(\pi)$ and advantage $A_t^{\text{on}}$, $\hat{V}_t = \sum_{l=0}^{\infty} \gamma^l R_{t+l}$, and $\lambda_V$ weights the value function loss. The overall three-stage training pipeline is summarized in Algorithm 1.

**Distillation to One-step Consistency Policy**

**High-frequency control** is crucial for robotics applications. While our diffusion policy achieves strong performance, the $K$-step denoising process introduces latency that can limit real-time deployment. To address this, we jointly train a consistency model $c_w$ that learns to directly map noise to actions in a single step, distilling knowledge from the multi-step diffusion teacher $\pi_\theta$.

During both offline and online RL training, we augment the policy optimization objective with the consistency distillation loss from Equation (S12):

$$\mathcal{L}_{\text{total}} = \mathcal{L}_{\text{RL}} + \lambda_{\text{CD}} \cdot \mathcal{L}_{\text{CD}}, \tag{11}$$

where $\mathcal{L}_{\text{RL}}$ is either the offline objective (Equation (8) with IQL-based advantages) or the online objective (Equation (10) with GAE). The consistency loss $\mathcal{L}_{\text{CD}}$ follows Equation (S12), with the teacher being our diffusion policy $\pi_\theta$ that performs $K$-step denoising conditioned on observation. The stop-gradient operator ensures the teacher policy continues to improve through RL objectives while simultaneously serving as a distillation target.

High-frequency control is essential for practical robot deployment in industrial settings. First, faster control loops directly translate to improved task completion efficiency—a robot operating at 20 (Hz) can execute the same trajectory in half the time compared to 10 (Hz) operation, significantly increasing throughput in factory automation, where cycle time directly impacts productivity. During inference, the consistency model generates actions in a single forward pass: $a^{\tau_0} = c_w(a^{\tau_K}, \tau_K | o)$, achieving $K\times$ speedup (e.g., from 100 (ms) to 10 (ms) latency) while preserving the performance of the diffusion policy.

## Statistical Analysis

We calculated the success rate of all experiments by computing the proportion of successful rollouts relative to the total number of trajectories. We recorded the wall-clock time as the average time taken by policies to complete the task across successful trajectories. The speedup ratio ($X \times$ faster) is computed by $X = \max(N, N_{\text{baseline}})/\min(N, N_{\text{baseline}})$ where $N$ is the quantity-of-interest.

## References and Notes:

1. J. Cui, J. Trinkle, Toward next-generation learned robot manipulation. *Sci. Robot.* **6**, eabd9461 (2021).
2. J. Luo, C. Xu, J. Wu, S. Levine, Precise and dexterous robotic manipulation via human-in-the-loop reinforcement learning. *Sci. Robot.* **10**, eads5033 (2025).
3. C. Chi, Z. Xu, S. Feng, E. Cousineau, Y. Du, B. Burchfiel, R. Tedrake, S. Song, Diffusion policy: Visuomotor policy learning via action diffusion. *Int. J. Robot. Res.* **44**, 1684-1704 (2024).
4. Y. Ze, G. Zhang, K. Zhang, C. Hu, M. Wang, H. Xu, "3D diffusion policy: Generalizable visuomotor policy learning via simple 3D representations" in *Proceedings of Robotics: Science and Systems* (RSS Foundation, 2024), 10.15607/RSS.2024.XX.067.
5. K. Black, N. Brown, D. Driess, A. Esmail, M. Equi, C. Finn, N. Fusai, L. Groom, K. Hausman, B. Ichter, S. Jakubczak, T. Jones, L. Ke, S. Levine, A. Li-Bell, M. Mothukuri, S. Nair, K. Pertsch, L. X. Shi, J. Tanner, Q. Vuong, A. Walling, H. Wang, U. Zhilinsky, $\pi_0$: a vision-language-action flow model for general robot control. arXiv: 2410.24164 [cs.LG] (2024).
6. K. Black, N. Brown, J. Darpinian, K. Dhabalia, D. Driess, A. Esmail, M. Equi, C. Finn, N. Fusai, M. Y. Galliker, D. Ghosh, L. Groom, K. Hausman, B. Ichter, S. Jakubczak, T. Jones, L. Ke, D. LeBlanc, S. Levine, A. Li-Bell, M. Mothukuri, S. Nair, K. Pertsch, A. Z. Ren, L. X. Shi, L. Smith, J. T. Springenberg, K. Stachowicz, J. Tanner, Q. Vuong, H. Walke, A. Walling, H. Wang, L. Yu, U. Zhilinsky, "$\pi_{0.5}$: a vision-language-action model with open-world generalization" in *Proceedings of the 9th Conference on Robot Learning* (PMLR, 2025), pp.17–40.
7. Z. Yuan, T. Wei, L. Gu, P. Hua, T. Liang, Y. Chen, H. Xu, HERMES: Human-to-robot embodied learning from multi-source motion data for mobile dexterous manipulation. arXiv: 2508.20085 [cs.RO] (2025).
8. T. Lin, K. Sachdev, L. Fan, J. Malik, Y. Zhu, "Sim-to-real reinforcement learning for vision-based dexterous manipulation on humanoids" in *Proceedings of the 9th Conference on Robot Learning* (PMLR, 2025), pp.4926–4940.
9. J. Luo, Z. Hu, C. Xu, Y. L. Tan, J. Berg, A. Sharma, S. Schaal, C. Finn, A. Gupta, S. Levine, "SERL: A Software Suite for Sample-Efficient Robotic Reinforcement Learning" in *2024 IEEE International Conference on Robotics and Automation (ICRA)* (IEEE, 2024), pp. 16961–16969.
10. L. Guo, Z. Xue, Z. Xu, H. Xu, "DemoSpeedup: Accelerating visuomotor policies via entropy-guided demonstration acceleration" in *Proceedings of the 9th Conference on Robot Learning* (PMLR, 2025), pp.599–609.
11. J. Song, C. Meng, S. Ermon, "Denoising diffusion implicit models" in *9th International Conference on Learning Representations (ICLR)* (ICLR, 2021), pp. 1-20.
12. Y. Song, P. Dhariwal, M. Chen, I. Sutskever, "Consistency Models" in *Proceedings of the 40th International Conference on Machine Learning (ICML)* (PMLR, 2023), pp. 32211–32252.
13. T. Z. Zhao, V. Kumar, S. Levine, C. Finn, "Learning fine-grained bimanual manipulation with low-cost hardware" in *Proceedings of Robotics: Science and Systems* (RSS Foundation, 2023), 10.15607/RSS.2023.XIX.016.
14. J. Ho, A. Jain, P. Abbeel, Denoising diffusion probabilistic models. in *Advances in Neural Information Processing Systems 33 (NeurIPS)* (Curran Associates, 2020), pp. 6840–6851.
15. Y. Song, J. Sohl-Dickstein, D. P. Kingma, A. Kumar, S. Ermon, B. Poole, "Score-based generative modeling through stochastic differential equations" in *9th International Conference on Learning Representations (ICLR)* (ICLR, 2021), pp. 1-36.
16. Y. Lipman, R. T. Chen, H. Ben-Hamu, M. Nickel, M. Le, "Flow matching for generative modeling" in *11th International Conference on Learning Representations* (ICLR, 2023), pp. 1-28.

17. Z. Wang, J. J. Hunt, M. Zhou, "Diffusion policies as an expressive policy class for offline reinforcement learning" in *11th International Conference on Learning Representations* (ICLR, 2023), pp. 1-17.
18. B. Kang, X. Ma, C. Du, T. Pang, S. Yan, "Efficient diffusion policies for offline reinforcement learning" in *Advances in Neural Information Processing Systems 36 (NeurIPS)* (Curran Associates, 2023), pp. 67195-67212.
19. C. Lu, H. Chen, J. Chen, H. Su, C. Li, J. Zhu, "Contrastive energy prediction for exact energy-guided diffusion sampling in offline reinforcement learning" in *Proceedings of the 40th International Conference on Machine Learning (ICML)* (PMLR, 2023), pp.22825-22855.
20. S. Ding, K. Hu, Z. Zhang, K. Ren, W. Zhang, J. Yu, J. Wang, Y. Shi, "Diffusion-based reinforcement learning via q-weighted variational policy optimization" in *Advances in Neural Information Processing Systems 37 (NeurIPS)* (Curran Associates, 2024), pp.53945-53968.
21. M. Psenka, A. Escontrela, P. Abbeel, Y. Ma, "Learning a diffusion model policy from rewards via q-score matching" in *Proceedings of the 41st International Conference on Machine Learning (ICML)* (PMLR, 2024), pp.41163-41182.
22. H. He, C. Bai, K. Xu, Z. Yang, W. Zhang, D. Wang, B. Zhao, X. Li, "Diffusion model is an effective planner and data synthesizer for multi-task reinforcement learning" in *Advances in Neural Information Processing Systems 36 (NeurIPS)* (Curran Associates, 2023), pp. 64896-64917.
23. Z. Ding, C. Jin, "Consistency models as a rich and efficient policy class for reinforcement learning" in *12th International Conference on Learning Representations (ICLR)* (ICLR, 2024), pp. 1-20.
24. H. Chen, C. Lu, Z. Wang, H. Su, J. Zhu, "Score regularized policy optimization through diffusion behavior" in *12th International Conference on Learning Representations (ICLR)* (ICLR, 2024), pp. 1-20.
25. H. Li, Z. Jiang, Y. CHEN, D. Zhao, "Generalizing consistency policy to visual RL with prioritized proximal experience regularization" in *Advances in Neural Information Processing Systems 37 (NeurIPS)* (Curran Associates, 2024), pp. 109672-109700.
26. K. Black, M. Janner, Y. Du, I. Kostrikov, S. Levine, "Training diffusion models with reinforcement learning" in *12th International Conference on Learning Representations (ICLR)* (ICLR, 2024), pp. 1-23.
27. Y. Fan, O. Watkins, Y. Du, H. Liu, M. Ryu, C. Boutilier, P. Abbeel, M. Ghavamzadeh, K. Lee, K. Lee, "DPOK: Reinforcement learning for fine-tuning text-to-image diffusion models" in *Advances in Neural Information Processing Systems 36 (NeurIPS)* (Curran Associates, 2023), pp.79858-79885.
28. A. Z. Ren, J. Lidard, L. L. Ankile, A. Simeonov, P. Agrawal, A. Majumdar, B. Burchfiel, H. Dai, M. Simchowitz, "Diffusion policy policy optimization" in *13th International Conference on Learning Representations (ICLR)* (ICLR, 2025), pp. 1-42.
29. S. Park, Q. Li, S. Levine, "Flow q-learning" in *Proceedings of the 42nd International Conference on Machine Learning (ICML)* (PMLR, 2025), pp. 48104-48127.
30. T, Nguyen, C. D. Yoo, Revisiting Diffusion Q-Learning: From Iterative Denoising to One-Step Action Generation. arXiv: 2508.13904 [cs.LG] (2025).
31. S. Mani, S. Venkataraman, A. Chandra, A. Rizvi, Y. Sirvi, S. Bhattacharya, A. Hazra, DiffClone: Enhanced behaviour cloning in robotics with diffusion-driven policy learning. arXiv: 2401.09243 [cs.RO] (2024).
32. S. F. Chen, H. C. Wang, M. H. Hsu, C. M. Lai, S. H. Sun, "Diffusion model-augmented behavioral cloning" in *Proceedings of the 41st International Conference on Machine Learning (ICML)* (PMLR, 2024), pp.7486-7510.

33. S. Nair, A. Rajeswaran, V. Kumar, C. Finn, A. Gupta, "R3M: A universal visual representation for robot manipulation" in *Proceedings of the 6th Conference on Robot Learning* (PMLR, 2023), pp.892-909.
34. A. Majumdar, K. Yadav, S. Arnaud, Y. Jason Ma, C. Chen, S. Silwal, A. Jain, V. P. Berges, T. Wu, J. Vakil, P. Abbeel, J. Malik, D. Batra, Y. Lin, O. Maksymets, A. Rajeswaran, F. Meier, "Where are we in the search for an artificial visual cortex for embodied intelligence?" in *Advances in Neural Information Processing Systems 36 (NeurIPS)* (Curran Associates, 2023), pp.655-677.
35. K. He, H. Fan, Y. Wu, S. Xie, R. Girshick, "Momentum contrast for unsupervised visual representation learning" in *2020 IEEE/CVF Conference on Computer Vision and Pattern Recognition (CVPR)* (IEEE, 2020), pp. 9726–9735.
36. Q. Zhang, Z. Liu, H. Fan, G. Liu, B. Zeng, S. Liu, "FlowPolicy: Enabling fast and robust 3D flow-based policy via consistency flow matching for robot manipulation" in *Proceedings of the 39-th AAAI Conference on Artificial Intelligence (AAAI)* (AAAI, 2025), pp.14754-14762
37. Y. Lu, Y. Tian, Z. Yuan, X. Wang, P. Hua, Z. Xue, H. Xu, H3DP: Triply-hierarchical diffusion policy for visuomotor learning. arXiv: 2505.07819 [cs.RO] (2025).
38. M. Janner, Y. Du, J. B. Tenenbaum, S. Levine, "Planning with diffusion for flexible behavior synthesis" in *Proceedings of the 39th International Conference on Machine Learning (ICML)* (PMLR, 2022), pp. 9902–9915.
39. A. Ajay, Y. Du, A. Gupta, J. Tenenbaum, T. Jaakkola, P. Agrawal, "Is conditional generative modeling all you need for decision making?" in *11th International Conference on Learning Representations (ICLR)* (ICLR, 2023), pp. 1-22.
40. S. Hegde, S. Das, G. Salhotra, G. S. Sukhatm. Latent weight diffusion: Generating reactive policies instead of trajectories. arXiv: 2410.14040 [cs.RO] (2024).
41. Z. Wang, M. Li, A. Mandlekar, Z. Xu, J. Fan, Y. Narang, L. Fan, Y. Zhu, Y. Balaji, M. Zhou, M. Y. Liu, Y. Zeng, "One-step diffusion policy: Fast visuomotor policies via diffusion distillation" in *Proceedings of the 42nd International Conference on Machine Learning (ICML)* (PMLR, 2025), pp. 63399-63416.
42. A. Bartsch, A. Car, A. B. Farimani, PinchBot: Long-horizon deformable manipulation with guided diffusion policy. ArXiv: 2507.17846 [cs.RO] (2025).
43. H. R. Walke, K. Black, T. Z. Zhao, Q. Vuong, C. Zheng, P. Hansen-Estruch, A. W. He, V. Myers, M. J. Kim, M. Du, A. Lee, K. Fang, C. Finn, S. Levine, "BridgeData V2: a dataset for robot learning at scale" in *Proceedings of the 7th Conference on Robot Learning* (PMLR, 2023), pp.1723-1736.
44. A. Kumar, A. Zhou, G. Tucker, S. Levine, "Conservative q-learning for offline reinforcement learning" in *Advances in Neural Information Processing Systems 33 (NeurIPS)* (Curran Associates, 2020), pp. 1179–1191.
45. X. B. Peng, A. Kumar, G. Zhang, S. Levine, Advantage-weighted regression: Simple and scalable off-policy reinforcement learning. ArXiv: 1910.00177 [cs.LG] (2019).
46. M. Nakamoto, Y. Zhai, A. Singh, M. S. Mark, Y. Ma, C. Finn, A. Kumar, S. Levine, "Cal-QL: Calibrated offline RL pre-training for efficient online fine-tuning" in *Advances in Neural Information Processing Systems 36 (NeurIPS)* (Curran Associates, 2023), pp.62244-62269.
47. K. Lei, Z. He, C. Lu, K. Hu, Y. Gao, H. Xu, "Uni-O4: Unifying online and offline deep reinforcement learning with multi-step on-policy optimization" in *12th International Conference on Learning Representations (ICLR)* (ICLR, 2024), pp. 1-34.
48. Z. Zhuang, K. LEI, J. Liu, D. Wang, Y. Guo, "Behavior proximal policy optimization" in *11th International Conference on Learning Representations (ICLR)* (ICLR, 2023), pp. 1-20.
49. J. Schulman, F. Wolski, P. Dhariwal, A. Radford, O. Klimov, Proximal policy optimization algorithms. ArXiv: 1707.06347 [cs.LG] (2017).

50. P. J. Ball, L. Smith, I. Kostrikov, S. Levine, "Efficient Online Reinforcement Learning with Offline Data" in *Proceedings of the 40th International Conference on Machine Learning (ICML)* (PMLR, 2023), pp. 1577–1594.
51. A. Nair, A. Gupta, M. Dalal, S. Levine, AWAC: Accelerating online reinforcement learning with offline datasets. ArXiv: 2006.09359 [cs.LG] (2020).
52. A. Wagenmaker, Y. Zhang, M. Nakamoto, S. Park, W. Yagoub, A. Nagabandi, A. Gupta, S. Levine, "Steering your diffusion policy with latent space reinforcement learning" in *Proceedings of the 9th Conference on Robot Learning* (PMLR, 2025), pp.258-282.
53. S. Levine, C. Finn, T. Darrell, P. Abbeel, End-to-end training of deep visuomotor policies. *J. Mach. Learn. Res.* **17**, 1334–1373 (2016).
54. D. Kalashnikov, A. Irpan, P. Pastor, J. Ibarz, A. Herzog, E. Jang, D. Quillen, E. Holly, M. Kalakrishnan, V. Vanhoucke, S. Levine, "Scalable deep reinforcement learning for vision-based robotic manipulation" in *Proceedings of the 2nd Conference on Robot Learning* (PMLR, 2018), pp.651-673.
55. T. Haarnoja, A. Zhou, P. Abbeel, S. Levine, "Soft actor-critic: Off-policy maximum entropy deep reinforcement learning with a stochastic actor" in *Proceedings of the 35th International Conference on Machine Learning (ICML)* (PMLR, 2018), pp. 1861–1870.
56. S. Fujimoto, H. Hoof, D. Meger, "Addressing function approximation error in actor-critic methods" in *Proceedings of the 35th International Conference on Machine Learning (ICML)* (PMLR, 2018), pp. 1587–1596.
57. K. Chua, R. Calandra, R. McAllister, S. Levine, "Deep reinforcement learning in a handful of trials using probabilistic dynamics models" in *Advances in Neural Information Processing Systems 31 (NeurIPS)* (Curran Associates, 2018), pp.4754-4765.
58. M. Janner, J. Fu, M. Zhang, S. Levine, "When to trust your model: Model-based policy optimization" in *Advances in Neural Information Processing Systems 32 (NeurIPS)* (Curran Associates, 2019), pp.12519-12530.
59. B. Eysenbach, S. Gu, J. Ibarz, S. Levine, "Leave no trace: Learning to reset for safe and autonomous reinforcement learning" in *6th International Conference on Learning Representations (ICLR)* (ICLR, 2018), pp. 1-18.
60. A. Gupta, J. Yu, T.Z. Zhao, V. Kumar, A. Rovinsky, K. Xu, T. Devlin, S. Levine, "Reset-free reinforcement learning via multi-task learning: Learning dexterous manipulation behaviors without human intervention" in *IEEE International Conference on Robotics and Automation (ICRA)* (IEEE, 2021), pp. 6664-6671.
61. A. Singh, L. Yang, K. Hartikainen, C. Finn, S. Levine, "End-to-end robotic reinforcement learning without reward engineering" in *Robotics: Science and Systems* (RSS Foundation, 2019), 10.15607/RSS.2019.XV.073.
62. P. F. Christiano, J. Leike, T. B. Brown, M. Martic, S. Legg, D. Amodei, "Deep reinforcement learning from human preferences" in *Advances in Neural Information Processing Systems 30 (NeurIPS)* (Curran Associates, 2017), pp.4299-4307.
63. T. Zhang, C. Yu, S. Su, Y. Wang, "ReinFlow: Fine-tuning flow matching policy with online reinforcement learning" in *Advances in Neural Information Processing Systems 38 (NeurIPS)* (Curran Associates, 2025), pp. 1-38.
64. A. Rajeswaran, V. Kumar, A. Gupta, G. Vezzani, J. Schulman, E. Todorov, S. Levine, "Learning complex dexterous manipulation with deep reinforcement learning and demonstrations" in *Proceedings of Robotics: Science and Systems* (RSS Foundation, 2018), 10.15607/RSS.2018.XIV.049.
65. T. Yu, D. Quillen, Z. He, R. Julian, K. Hausman, C. Finn, S. Levine, "Meta-world: A benchmark and evaluation for multi-task and meta reinforcement learning" in *Conference on Robot Learning* (PMLR, 2020), pp. 1094–1100.

66. J. Fu, A. Kumar, O. Nachum, G. Tucker, S. Levine, D4RL: Datasets for Deep Data-Driven Reinforcement Learning. ArXiv: 2004.07219 [cs.LG] (2020).
67. E. Todorov, T. Erez, Y. Tassa, "MuJoCo: A physics engine for model-based control" in *2012 IEEE/RSJ International Conference on Intelligent Robots and Systems* (IEEE, 2012), pp. 5026–5033.
68. I. Kostrikov, A. Nair, S. Levine, "Offline reinforcement learning with implicit q-learning" in *10th International Conference on Learning Representations (ICLR)* (ICLR, 2022), pp. 1-11.
69. J. Schulman, P. Moritz, S. Levine, M. I. Jordan, P. Abbeel, "High-dimensional continuous control using generalized advantage estimation" in *4th International Conference on Learning Representations (ICLR)* (ICLR, 2016), pp. 1-14.

**Acknowledgments:**

We would like to thank Tianhai Liang for his contributions to the hardware design for this project, and Jiacheng You and Zhecheng Yuan for their valuable discussions and insights.

**Funding:** TBD

**Author contributions:** **K.L.**: Project lead, Core, responsible for the original idea and codebase, overall design and training of simulation and real-world experiments, and paper writing. **H.L.**: Core, optimized training infrastructure, led the unscrew, pour and juicing tasks, contributed to paper writing, and implemented part of the simulation baseline. **D.Y.**: Core, led the juicing task, contributed to paper writing, most of the real-world baselines, and implemented part of the simulation baseline. **Z.W.**: Core, responsible for the real-world robot setup, design, and data collection, and contributed to paper writing. **L.G.**: Core, led data collection for the folding task, managed its offline training process, and explored dual-arm embodiments in the early stages. **Z.J.**: Core, contributed to part of the algorithm design, improved consistency policy distillation, debugged and explored settings in simulation, and developed most of the simulation baselines. **Z.W.**: Managed the Metaworld domain tasks. **S.L.**: Writing refinement. **H.X.**: Principal Investigator (PI), Core, responsible for project direction and guidance, and contributed to paper writing.

**Competing interests:** TBD.

**Data and materials availability:** Additional material can be found at https://lei-kun.github.io/RL-100/

**Supplementary Materials:**

**The file includes:**

Tables S1 to S3.

Figs. S1 to S8.

Description of Tasks

Preliminaries (Equation S1 to S14).

Simulation Results and Analysis.

Supplementary Experiment Details.

Additional Related Work.

**Other Supplementary Material for this manuscript includes the following:**

Movies S1–S12 and R1-R6 (Movie list):

Movie R1: Extreme OOD test of Orange Juicing.

Movie R2: Extreme OOD test of Pouring.

Movie R3: Teaser video. One real-time (1x) trajectory of Box Folding.

Movie.R4: Failure cases of baselines, and consecutive successful trials on Box Folding.

Movie.R5: Extreme OOD test of Box Folding.

Movie.R6: Robustness to human disturbances on Box Folding.

Movie S1. Teaser video. One real-time (1x) trajectory per task.

Movie S2. Ablations. Zero-shot adaptation; few-shot adaptation; robustness to disturbances.

Movie S3. Execution efficiency comparison. Time-to-completion versus baselines.

Movie S4. Human–robot comparison. Human & Robot competition on Bowling and Push-T.

Movie S5. Real-world deployment. Zero-shot deployment of Juicing task in a shopping mall.

Movie S6. Dynamic Push-T. Consecutive successful trials on Dynamic Push-T.

Movie S7. Agile Bowling. Consecutive successful trials on Agile Bowling.

Movie S8. Pouring. Consecutive successful trials on Pouring.

Movie S9. Dynamic Unscrewing. Consecutive successful trials on Dynamic Unscrewing.

Movie S10. Soft-towel Folding. Consecutive successful trials on Soft-towel Folding.

Movie S11. Orange Juicing—Placing. Consecutive successful trials on Juicing—Placing.

Movie S12. Orange Juicing—Removal. Consecutive successful trials on Juicing—Removal.

# Supplementary Materials for

**RL-100: Performant Robotic Manipulation with Real-World Reinforcement Learning**

Kun Lei *et al.*

Corresponding author: Huazhe Xu, huazhe_xu@mail.tsinghua.edu.cn

**The file includes:**

Tables S1 to S3.
Figs. S1 to S8.
Description of Tasks
Preliminaries (Equation S1 to S14).
Simulation Results and Analysis.
Supplementary Experiment Details.
Additional Related Work.

**Other Supplementary Material for this manuscript includes the following:**

Movies S1–S12 and R1-R6 (Movie list):
Movie R1: Extreme OOD test of Orange Juicing.
Movie R2: Extreme OOD test of Pouring.
Movie R3: Teaser video. One real-time (1x) trajectory of Box Folding.
Movie.R4: Failure cases of baselines, and consecutive successful trials on Box Folding.
Movie.R5: Extreme OOD test of Box Folding.
Movie.R6: Robustness to human disturbances on Box Folding.
Movie S1. Teaser video. One real-time (1x) trajectory per task.
Movie S2. Ablations. Zero-shot adaptation; few-shot adaptation; robustness to disturbances.
Movie S3. Execution efficiency comparison. Time-to-completion versus baselines.
Movie S4. Human–robot comparison. Human & Robot competition on Bowling and Push-T.
Movie S5. Real-world deployment. Zero-shot deployment of Juicing task in a shopping mall.
Movie S6. Dynamic Push-T. Consecutive successful trials on Dynamic Push-T.
Movie S7. Agile Bowling. Consecutive successful trials on Agile Bowling.
Movie S8. Pouring. Consecutive successful trials on Pouring.
Movie S9. Dynamic Unscrewing. Consecutive successful trials on Dynamic Unscrewing.
Movie S10. Soft-towel Folding. Consecutive successful trials on Soft-towel Folding.
Movie S11. Orange Juicing—Placing. Consecutive successful trials on Juicing—Placing.
Movie S12. Orange Juicing—Removal. Consecutive successful trials on Juicing—Removal.

**Table S1. Task suite, embodiments, and control modes.** Embodiments listed are typical examples; our framework is task- and embodiment-agnostic. Control modes are selected per task as either single-step (one action per control tick) or action-chunk (short $c$-step segments).

| Task | Control mode | Embodiments | Modality | Key challenges |
|---|---|---|---|---|
| Dynamic Push-T | Single-step | *UR5, 3D-printed end-effector (single-arm)* | Rigid-body dynamics | Fast reaction to a moving goal pose and online perturbations |
| Agile Bowling | Single-step | *UR5, 3D-printed end-effector (single-arm)* | Rigid-body dynamics | Release-timing control at high velocity; trajectory and release-pose accuracy |
| Pouring | Single-step | *Franka, LeapHand (single-arm)* | Granular | Spillage minimization; flow control; container alignment under motion |
| Dynamic Unscrewing | Action-chunk | *Franka, LeapHand (single-arm)* | Precision assembly | Time-varying alignment; torque/pose regulation; cross-thread avoidance |
| Soft-towel Folding | Action-chunk | *xArm, Franka, Robotiq (dual-arm)* | Deformable object | Large deformation; coordinated contacts; fold accuracy |
| Orange Juicing[‡] | Action-chunk | *xArm, Robotiq (single-arm)* | Deformable object | Confined-space insertion/ejection; generalization to fruit variability |
| Box Folding | Action-chunk | *UR5, 3D-printed end-effector (dual-arm)* | Deformable object (paperboard) | Long-horizon bimanual coordination with error-sensitive crease/edge alignment; complex contact-rich transitions during folding with a semi-rigid deformable object; Spatial generalization unseen initial shapes and orientations |

Control modes: Single-step (one action per control tick); Action-chunk (short $c$-step segments). [‡]Two subtasks: *Placing* (place fruit into the press zone) and *Removal* (remove the spent fruit); the spent fruit is *deformable*, fruit sizes vary substantially (requiring strong generalization), and operation occurs in a *confined* cavity with narrow clearances.

**Table S2. State and action components of each task.** We abbreviate dimension as 'dim.', proprioception as 'prop.'.

| Task | Observation space | | | Action space | |
|---|---|---|---|---|---|
| | Point cloud dim. | Prop. dim. | Note | Dim. | Note |
| Dynamic Push-T | (512, 3) | 6 | UR5 joints | 2 | Normalized $\Delta(x, y)$ of end-effector |
| Agile Bowling | (512, 3) | 6 | Same as above | 2 | Same as above |
| Pouring | (512, 3) | 23 = 7 + 16 | Franka joints, LeapHand joints | 22 = 6 + 16 | Target arm pose, target hand joints |
| Dynamic Unscrewing | (1024, 3) | 23 | Same as above | 23 = 22 + 1 | 22: Same as above, 1: enable arm motion |
| Soft-towel Folding | (1024, 3) | 15 = 7+8 | Dual arm joints, grippers state | 16 = 2 × (7+1) | Dual arms: target pose (pos+quat), target gripper |
| Orange Juicing — Placing | (1024, 3) | 7 = 6 + 1 | xArm joints, gripper state | 8 = 6 + 1 + 1 | Target arm pose, target gripper, task done |
| Orange Juicing — Removal | (1024, 3) | 7 | Same as above | 8 | Same as above |
| Box Folding | (1024, 3) | 14 = 2 × (6+1) | Dual arm joints, grippers state | 14 = 2 × (6+1) | Dual arms: target pose, target gripper |

**Table S3. Summary of numbers of episodes and corresponding collection time (h) across all tasks during the three training stages - human demonstration, iterative offline RL, and online RL.** Note that in offline RL stage, the number is the sum of all iterations.

| Task | Human Demonstration | | Iterative Offline RL | | Online RL | |
|---|---|---|---|---|---|---|
| | # of epi. | Collection time (h) | # of epi. | Collection time (h) | # of epi. | Collection time (h) |
| Dynamic Push-T | 100 | 2 | 821 | 8 | 763 | 7.5 |
| Agile Bowling | 100 | 2 | 249 | 2 | 213 | 2.5 |
| Pouring | 64 | 1 | 741 | 6.8 | 129 | 1.5 |
| Soft-towel Folding | 400 | 5 | 896 | 11 | 654 | 8.5 |
| Dynamic Unscrew | 31 | 0.5 | 467 | 4.5 | 288 | 3 |
| Orange Juicing-Placing | 80 | 1.5 | 642 | 10.5 | 750 | 12.5 |
| Orange Juicing-Removal | 29 | 0.5 | 149 | 2.5 | 240 | 4 |
| Box Folding | 200 | 3 | 608 | 9.3 | 186 | 3 |
| **Average** | **126** | **1.9** | **572** | **6.8** | **403** | **5.3** |

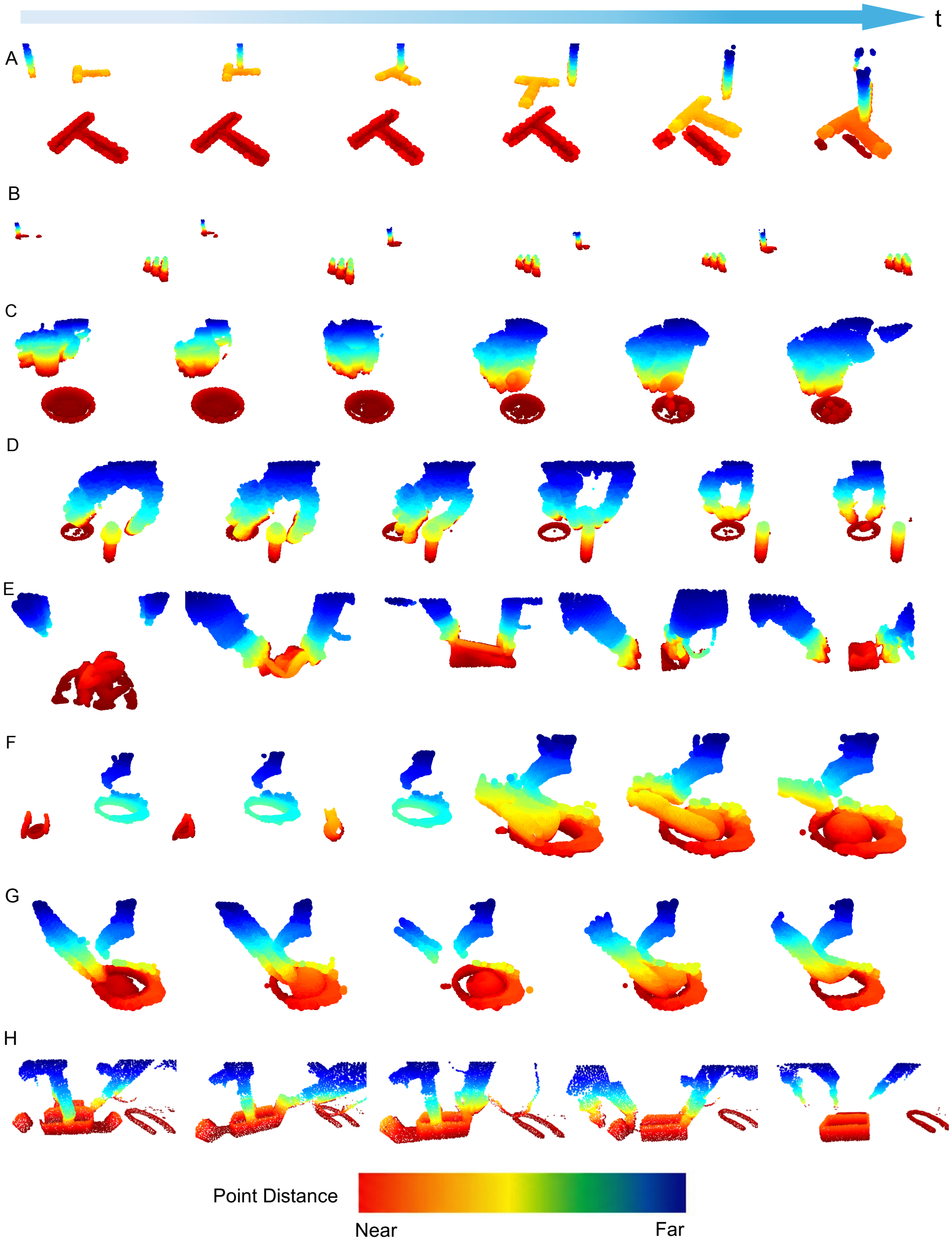

**Fig. S1. Point-cloud observation trajectories used by the policy during deployment.** Each row shows time-ordered point-cloud observations (left→right, see arrow) in the camera frame for one task, after cropping and down-sampling (farthest-point sampling). Colors indicate depth for visualization only (red = near, blue = far; see depth legend) and are not used by the policy. Tasks are: (A) Push-T, (B) Bowling, (C) Pouring, (D) Unscrewing, (E) Folding, (F) Orange Juicing—Placing, (G) Orange Juicing—Removal, (H) Box Folding. This figure highlights: (i) partial observability of raw 3D perception (occlusions and view-dependent visibility); and (ii) task-dependent temporal geometry changes over the rollout (rigid-body motion, contact interactions, deformable-object changes), motivating point clouds (together with proprioception) as a robust observation modality for closed-loop control.

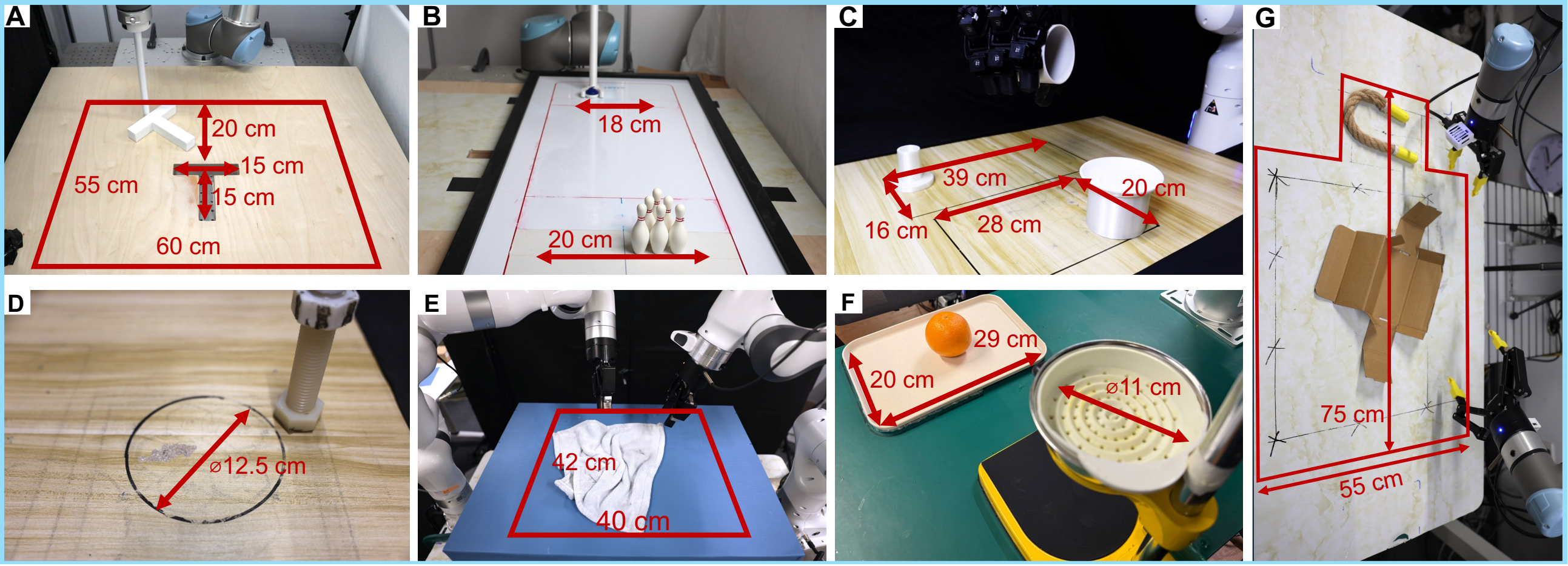

**Fig. S2. The object initialization workspaces for the eight real-world tasks.** At the beginning of each episode, objects-of-interest are randomly positioned within the areas denoted by the red boundaries and annotations.

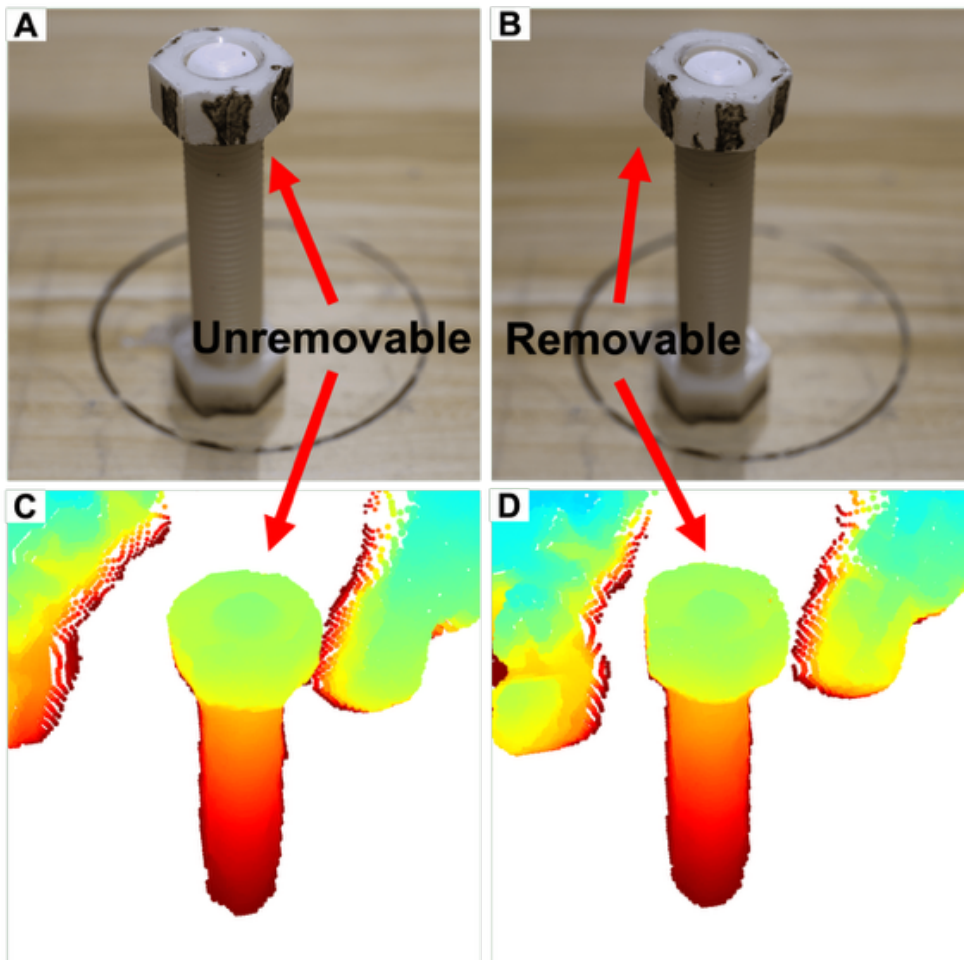

**Fig. S3. Comparison between different stages of unscrewing task.** (A) an unremovable nut and (B) a removable one, as well as their corresponding point-cloud observations (C-D). Robot policies must be accurate enough to recognize the removable state by the tiny tilt shown in (B) and (D) to guarantee success grasps.

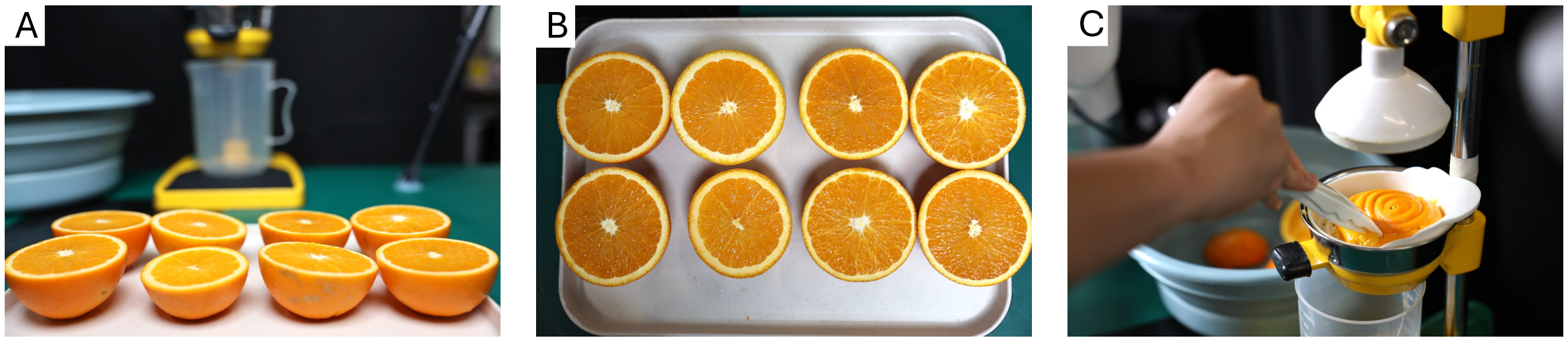

**Fig. S4. Challenging parts of the orange juicing task.** (A)-(B) spatial robustness w.r.t. orange appearances and positions; (C) force-sensitive manipulation of deformable orange discards.

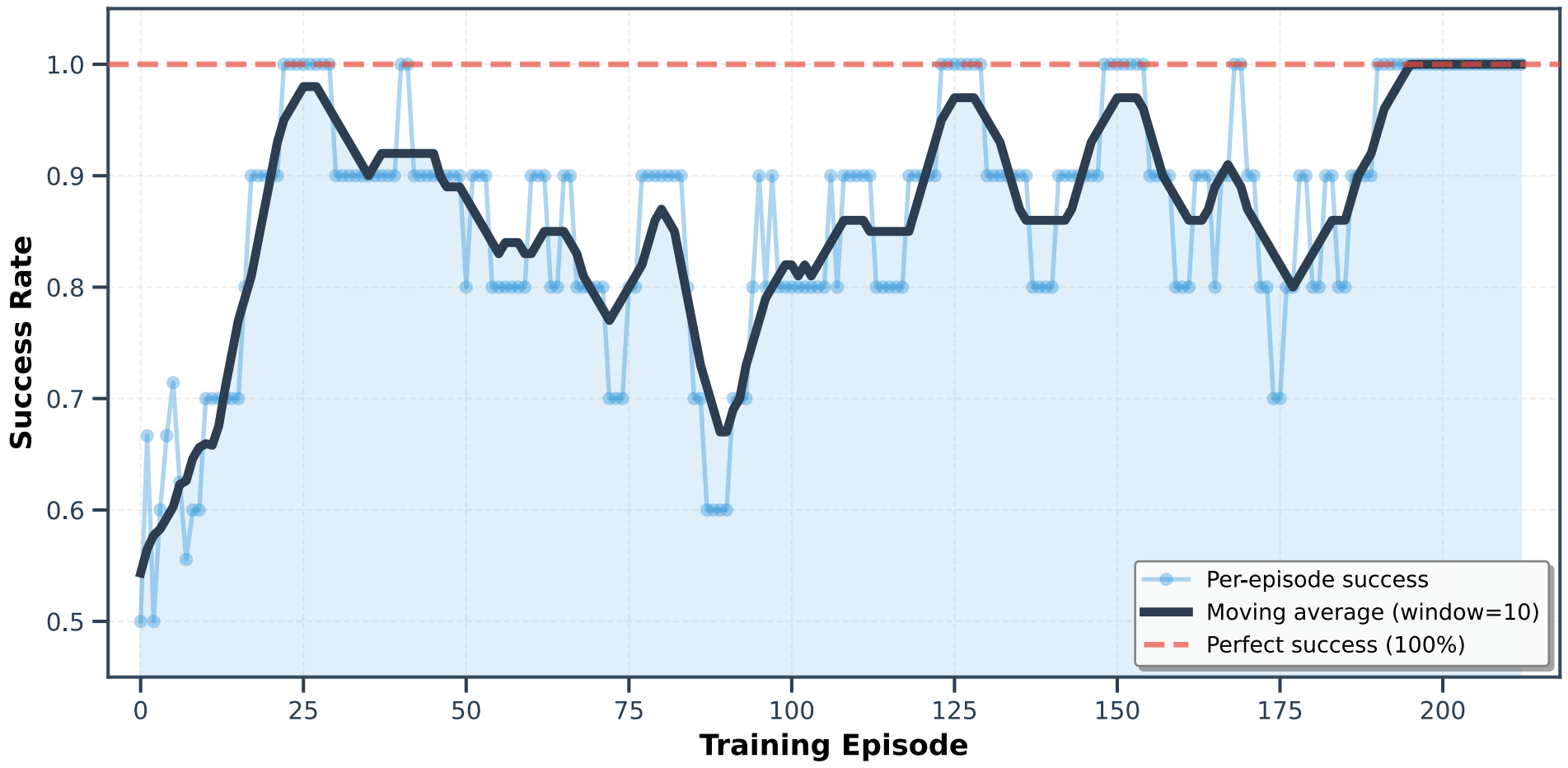

**Fig. S5. Online RL training curve for Agile Bowling.** Per-episode success rate during online RL finetuning phase, showing rapid convergence from the iterative offline baseline (approximately 80+%) to perfect performance (100%).

# Description of Tasks

Following the overview, we describe the objective, challenges, and further evaluation protocols beyond the normal ones of each task individually.

### Dynamic Push-T

The Dynamic Push-T task requires the robot arm, equipped with a 3D-printed stick-like end-effector, to push a T-shaped block from various initial locations into a wooden slot that is only slightly larger (3mm each side) than the block. The T-block's initial position and orientation are fully randomized, uniformly distributed within a 55cm × 60cm reachable workspace of the arm. The robot's initial position and the target slot's position are fixed. The task is considered ***successful*** only when the T-block is fully and correctly inserted into the slot.

As a 3D extension of the 2D Push-T task, this task inherits the challenge of requiring high-frequency and precise dynamic adjustments but introduces ***additional complexities***. For instance, due to the slot's geometry, the friction coefficient at the slot's edges varies when the T-block is partially suspended, leading to unpredictable behaviors such as significant rotations of the block. Additionally, the robot must avoid pushing the T-block into the slot in an incorrect orientation, which could lead to failure and end-effector fracture. The policy must therefore exhibit robust control to handle these dynamic and frictional variations while ensuring precise alignment with the slot.

To evaluate the generalization capabilities of the framework, we further perform the task under multiple variations: (i) different initial positions and orientations of the T-block, (ii) a wooden surface with a desktop sticker to reduce friction, (iii) the presence of additional objects on the workspace to introduce visual input distractions, and (iv) external disturbances applied to the T-block's during pushing. These variations challenge the policy to dynamically adjust with high precision, ignore environmental distractions, and avoid incorrect paths, demonstrating its robustness in maintaining high-frequency dynamic control while accurately completing the task.

**Agile Bowling**

The Agile Bowling task requires the robot arm, equipped with a 3D-printed semi-circular end effector, to push a curling stone to knock down six bowling pins positioned 60cm away. The robot arm starts from a fixed initial position, while the curling stone and bowling pins are uniformly distributed along an 18cm starting line and a 20cm target line, respectively. The task is considered ***successful*** only when more than five pins are knocked down.

This task presents several ***challenges*** due to the highly sensitive dynamics of the setup. The curling surface is extremely smooth, and the end effector is 4mm larger in diameter than the curling stone, meaning even minor variations in the robot's movements can significantly alter the stone's trajectory. The robot must precisely initiate the stone's motion from its starting position, align it toward the bowling pins, and execute a controlled push while making fine adjustments to account for potential trajectory deviations. Furthermore, knocking down almost all pins requires applying sufficient force to accurately strike the first pin, as slight misalignments can result in some pins remaining upright, leading to failure. The small size of both the curling stone and bowling pins provides limited visual input, adding to the challenge of achieving precise control.

To evaluate the generalization capabilities of the framework, we test the task under varying initial positions of the curling stone and bowling pins along their respective lines. We further test (i) zero-shot transferability on a coarser trial surface and (ii) adaptation by further finetuning on an inversed pins placement of the policy. These variations challenge the policy to adapt its pushing strategy and trajectory corrections to different configurations while maintaining accuracy.

**Pouring**

This task requires the robot to grasp a cup containing mixed nuts and snacks of varying sizes and textures with a Leap Hand dexterous hand, then precisely pour the contents into a target plate through controlled wrist rotation. The cup's initial position is randomized within 39cm×16cm on the table surface, while the plate position varies within 28cm×20cm. The robot begins from a fixed initial pose, grasps the cup, rotates the wrist to invert the container, and pours the contents into the plate.

***Success*** requires a stable grasp and accurate alignment with no misalignment-caused spillage (incidental bounce-outs ignored). ***Challenges*** include: (i) a smooth cup surface demands finger configurations that balance grip stability and pour orientation; (ii) large start-position randomization requires positional generalization for both grasp and pour trajectory; (iii) heterogeneous contents induce stochastic flow, calling for adaptive control to complete transfer without overshoot.

To test generalization, we add three variants: pouring soft small candies, water from a cup, and water via a long-spout teapot, probing transfer from granular to liquid materials and to altered container geometry. These variations test the policy's ability to transfer learned manipulation skills across different materials (granular solids to liquids) and container morphologies while maintaining the core competency of controlled pouring.

### Dynamic Unscrewing

This task simulates industrial assembly: a Leap Hand must unscrew a nut from a vertically mounted bolt and place it on a plate. The bolt location is randomized within a 12.5 cm-diameter circle on the work surface, demanding robust adaptation to varying target locations. The task involves a multi-phase manipulation sequence: the robot first approaches and uses its thumb to engage the nut with rotational motions, gradually lifting it along the threaded shaft. As the nut ascends, the robot must continuously adjust its hand pose to maintain effective contact and torque transmission. Once the nut clears the threads, the robot should precisely grasp it with a pinch grip between the index finger and thumb, then transports and places it on the plate.

***Success*** requires (i) complete unscrewing, (ii) a stable grasp without drops, and (iii) accurate placement. ***Challenges*** include precise contact initiation for leverage, adaptive control as optimal contact points and force directions change during ascent, timely detection of thread clearance and rapid reconfiguration from rolling contact to pinch grip (see Fig. S3), and fine motor control for a small object. To test robustness, we introduce human-applied counter-rotational perturbations during unscrewing.

### Soft-towel Folding

This task requires dual robot arms to collaboratively fold a randomly crumpled towel, uniformly distributed in the 42cm×40cm central region of a table, into a neatly folded state. The task is considered ***successful*** only when the arms lift and flatten the towel, then perform two precise folds, ensuring no wrinkles or misalignments occur.

This task is ***challenging*** due to the need for precise dual-arm coordination and handling a deformable object with significant dynamic variability. The towel's initial position and crumpled configuration vary greatly, leading to substantial differences in point cloud observations. The task is further complicated by potential issues such as uneven folding, unflipped towel corners, or failed grasps, requiring the policy to perform robust failure recovery. Arms must dynamically adjust their motions to flatten and fold the towel accurately while responding to unexpected deformations or missteps. We make a simplification that the two corners for bimanual picking are always visible after initialization. However, this does not trivialize the task, since the main challenges remain long-horizon bimanual coordination, contact-rich manipulation, and controlling deformable cloth dynamics under diverse shapes.

To evaluate the generalization capabilities of the framework, we test the policy under conditions with external human-induced disturbances during the folding process, assessing its ability to recover from failures and adapt to disruptions. This task challenges the model's capacity for dual-arm coordination, generalization across diverse initial towel configurations, manipulation of deformable objects, and effective error correction.

**Orange Juicing**

The orange-juicing task comprises three subtasks: placing a halved orange on the juicer, pressing the lever to extract juice, and removing the discarded pulp. The robot's end effector is fitted with two 3D-printed sleeves that secure rubber grippers, enabling robust grasps on irregular fruit surfaces. For each subtask, the robot arm's initial and final poses are fixed; these poses are defined per subtask and therefore differ across the three subtasks.

***Orange placing.*** A randomly sized half-orange is placed at a random position and orientation within a 29cm×20cm inner area of a tray. The robot first grasps the orange with the gripper oriented vertically to the tray surface, then rotates the wrist joint by 180○ to flip the orange, and finally places it onto the filter section of a lever-style juicer before retracting. ***Success*** is defined as the orange being stably seated on top of the filter. This step is ***challenging*** due to large variability in size, inclination, height, and placement (see Fig. S4A and Fig. S4B), requiring strong robustness to spatial randomness.

***Lever pressing.*** The pressing motion follows a fixed trajectory. To ensure consistent execution, we record the downward and upward lever motions via kinesthetic teaching. At deployment, the robot replays this trajectory to lower the press, squeeze the juice, and reset the lever. ***Success*** is defined as fully depressing and restoring the lever.

***Discard removal.*** After pressing, the flattened orange half is randomly embedded within the 11cm diameter juicer filter. The robot begins with the gripper closed and aligned parallel to the filter surface, pushing against the flattened fruit to move it from the tightly embedded position. The wrist then rotates by 90○, the gripper opens to grasp the orange, and the discarded pulp is placed into a disposal container on the left side of the juicer. The task is considered ***successful*** if the flattened orange is removed from the filter and dropped into the container. This subtask is particularly ***challenging*** because it involves force-sensitive manipulation of a deformable object in a confined space. Fig. S4C indicates that sufficient force is needed to move the orange, but excessive force can cause deformation and hinder grasping. Furthermore, juice on the surface makes the orange slippery, and inappropriate grasping poses may result in failed pickups.

**Box Folding**

This task requires two robot arms to fold a pre-scored paperboard blank into a self-supporting box. At the beginning of each episode, the blank is placed flat on the tabletop with randomized in-plane translation and rotation within the 75cm×55cm central region of a table, and the robot starts from a fixed initial pose. The policy must establish stable grasps on thin, compliant edges, reorient the blank when needed, and execute a multi-stage folding sequence that brings panels up along the pre-defined creases and forms a closed 3D structure.

***Success*** requires forming the intended box geometry (all panels folded along the correct creases) and maintaining structural integrity without collapse or irreversible misfolds. The task is ***challenging*** due to: (i) long-horizon error compounding, where small geometric misalignments early in the sequence can cascade into irreversible failures; (ii) contact-rich, discontinuous interactions when folding a semi-rigid deformable object, including slip, spring-back, and intermittent contact changes along edges and creases; and (iii) strict crease/edge alignment and bimanual coordination, which often demands regrasping and substantial SE(3) reorientation to keep folds registered and avoid collisions/interference between hands and panels.

To evaluate spatial generalization and adaptation, we additionally test unseen initial shapes and orientations. In particular, we include an extreme OOD initialization regime with strong misalignment ($\theta \in [60°, 90°]$), where $\theta$ is the in-plane angle between the blank's principal axis and the bimanual manipulation axis, whereas training demonstrations are concentrated in near-aligned configurations (typically $\theta \in [-30°, 30°]$), as shown in Fig. 3F. This setting probes whether the policy can reorient under severe geometric distribution shift rather than relying on aligned starts. Moreover, we evaluate robustness under human disturbances during execution, including dragging, pushing, and tapping, as shown in Fig. 3N–Q.

# PRELIMINARIES

## Reinforcement Learning

We formulate the robotic manipulation problem as a Markov Decision Process (MDP) $\langle \mathcal{S}, \mathcal{A}, P, R, \gamma \rangle$, where $\mathcal{S}$ is the state space, $\mathcal{A}$ is the action space, $P$ is the transition dynamics, $R$ is the reward function, and $\gamma$ is the discount factor. The robot policy $\pi$ chooses action $a_t$ at state $s_t$ to maximize discounted cumulative rewards $\Sigma_{t=0}^{\infty} \gamma^t R(s_t, a_t)$. The value function $V(s) = \Sigma_{t=0}^{\infty} \gamma^t R(s_t, a_t)|_{s_0=s, a_t \sim \pi(\cdot|s_t)}$ is defined to measure robot performance starting from a given state $s$ and the Q function $Q(s,a) = \Sigma_{t=0}^{\infty} \gamma^t R(s_t, a_t)|_{s_0=s, a_0=a, a_t \sim \pi(\cdot|s_t)}$ starts from given $s$ and $a$.

**Offline-to-online RL.** Our post-training follows an offline-to-online paradigm. Following (*47*), we employ a *Proximal Policy Optimization (PPO)* objective (*48, 49*) to unify both stages. The core learning objective incorporates importance sampling for proximal policy updates:

$$J_i(\pi) = \mathbb{E}_{s \sim \rho_\pi, a \sim \pi_i}\left[\min\left(r(\pi)A(s,a), clip(r(\pi), 1-\epsilon, 1+\epsilon)A(s,a)\right)\right] \tag{S12}$$

where $\rho_\pi$ is the stationary state distribution induced by policy $\pi$, $r(\pi) = \pi(a|s)/\pi_i(a|s)$ is the importance ratio, and $A(s,a) = Q(s,a) - V(s)$ is the advantage function. The key distinction between offline and online stages lies in advantage estimation:

- **Offline:** Implicit Q Learning (IQL) (*68*) value functions: $A^{\text{off}}(s,a) = Q(s,a) - V(s)$;
- **Online:** Generalized Advantage Estimation (GAE) (*69*) $A^{\text{on}}(s,a) = \text{GAE}(R_t, V)$ to balance variance and bias.

## Diffusion Models

We overload the subscript $t$ from timestep in MDP to the step index in diffusion process in the following two subsections. Diffusion models (*14*) learn to reverse a noising process that gradually corrupts clean data $x_0 \in \mathbb{R}^d$ into Gaussian noise to reconstruct the original clean data distribution. Given a schedule $\{\alpha_t\}_{t=1}^T$ with $\alpha_t = 1 - \beta_t$, a sample $x_0$ drawn from the clean distribution, the forward noising process follows a closed form:

$$x_t = \sqrt{\overline{\alpha}_t} x_0 + \sqrt{1 - \overline{\alpha}_t} \varepsilon, \varepsilon \sim \mathcal{N}(0, \boldsymbol{I}), \tag{S13}$$

$$\overline{\alpha}_t = \prod_{s=1}^{t} \alpha_s. \tag{S14}$$

A denoiser $\varepsilon_\theta$ is trained to recognize the noise inside the noisy sample via:

$$\mathcal{L}_{\text{simple}}(\theta) = \mathbb{E}_{x_0,t,\varepsilon}[\|\varepsilon - \varepsilon_\theta(x_t, t)\|_2^2]. \tag{S15}$$

to recover the clean sample.

## DDIM Sampling with Stochastic Form

Denoising Diffusion Implicit Models (DDIM) (*11*) provide a family of samplers that interpolate between deterministic and stochastic generation. Given a learned denoiser $\varepsilon_\theta$, the predicted clean sample at time $t$ is commonly written as:

$$\hat{x}_0(x_t, t) = \frac{x_t - \sqrt{1 - \overline{\alpha}_t} \varepsilon_\theta(x_t, t)}{\sqrt{\overline{\alpha}_t}}, \tag{S16}$$

where $\overline{\alpha}_t = \prod_{i=1}^{t} \alpha_i$ denotes the cumulative noise schedule.

We consider a (possibly) coarse time-subsequence for sampling $\tau_K > \tau_{K-1} > \cdots > \tau_1$, with $K \ll T$ (for example, $T = 50\sim1000$ and $K = 5\sim10$). To cover both the single-step ($t \to t - 1$) and subsampled ($\tau_k \to \tau_{k-1}$) cases in a unified notation, denote a generic transition from time $t$ to an earlier time $m$ (with $m < t$) by $t \to m$. DDIM then defines a stochastic update with variance parameter $\sigma_{t\to m} \geq 0$ as:

$$\mu_\theta(x_t, t \to m) = \sqrt{\overline{\alpha}_m} \hat{x}_0(x_t, t) + \sqrt{1 - \overline{\alpha}_m - \sigma_{t\to m}^2} \varepsilon_\theta(x_t, t), \tag{S17}$$

$$x_m = \mu_\theta(x_t, t \to m) + \sigma_{t\to m} \varepsilon_{t\to m}, \varepsilon_{t\to m} \sim \mathcal{N}(0, \mathbf{I}). \tag{S18}$$

The radical in Equation (S17) requires the constraint:

$$1 - \overline{\alpha}_m - \sigma_{t\to m}^2 \geq 0, \text{hence } 0 \leq \sigma_{t\to m} \leq \sqrt{1 - \overline{\alpha}_m}. \tag{S19}$$

In particular, the deterministic DDIM update is recovered when $\sigma_{t\to m} = 0$ (the distribution degenerates to a Dirac at $\mu_\theta$). Conversely, positive values of $\sigma_{t\to m}$ inject stochasticity into the transition.

**Policy perspective and log-likelihood.** When $\sigma_{t\to m} > 0$ the transition from $x_t$ to $x_m$ can be viewed as a Gaussian sub-policy

$$\pi_\theta(x_m \mid x_t, t \to m) = \mathcal{N}(\mu_\theta(x_t, t \to m), \sigma_{t\to m}^2 \mathbf{I}), \tag{S20}$$

$$\log \pi_\theta( x_m \mid x_t, t \to m ) = -\frac{1}{2\sigma_{t\to m}^2} \|x_m - \mu_\theta(x_t, t \to m)\|^2 + C, \tag{S21}$$

where $C$ is a constant independent of the parameters $\theta$. Note that Equation (S21) is only valid for $\sigma_{t\to m} > 0$; when $\sigma_{t\to m} = 0$ the density becomes singular and the transition is best described as the deterministic mapping $x_m = \mu_\theta(x_t, t \to m)$ or equivalently a Dirac distribution.

A full DDIM sampling process recovers a clean sample $x_0$ by chaining the sub-policies $\{\pi_\theta(x_{\tau_{k-1}}|x_{\tau_k}, \tau_k \to \tau_{k-1})\}_{k=K}^{1}$, starting from $x_{\tau_K} \sim \mathcal{N}(0, \mathbf{I})$. In practice, one may set $\sigma_{t\to m} = 0$ for fully deterministic sampling or choose a small positive $\sigma_{t\to m}$ (subject to Equation (S19)) to tradeoff between sample diversity and stability. If the log-likelihood in Equation (S21) is later used as an objective or as part of a fine-tuning criterion, care must be taken in handling the $\sigma_{t\to m} \to 0$ limit (e.g., by restricting likelihood-based terms to steps with strictly positive variance).

**Notation and scheduling conventions.** Throughout the paper we distinguish MDP timesteps from diffusion (denoising) timesteps. Environmental timesteps are denoted by $t$, while diffusion indices follow a (possibly subsampled) schedule $\tau_K > \tau_{K-1} > \cdots > \tau_1$ , and are written as superscripts (e.g. $a^{\tau_k}$). A generic denoising transition from time $t$ to an earlier time $m$ is denoted by $t \to m$ (for the subsampled schedule this will typically be written $\tau_k \to \tau_{k-1}$ . Variance parameters are indexed consistently as $\sigma_{\tau_k \to \tau_{k-1}}$ (abbreviated as $\sigma_{\tau_k}$ when unambiguous). We always enforce the constraint:

$$0 \le \sigma_{\tau_k} \le \sqrt{1 - \bar{\alpha}_{\tau_{k-1}}}, \tag{S22}$$

so that all square roots appearing in the DDIM updates are real. When $\sigma_{\tau_k} = 0$ the corresponding transition degenerates to a deterministic mapping (Dirac), and Gaussian log-densities are not defined; therefore, any likelihood-based objective (e.g., policy-gradient using $\log\pi$) must only use steps with strictly positive variance.

## Consistency Models

Consistency models (*12*) learn a single-step mapping from noisy inputs at arbitrary noise levels to clean data. Denote the consistency model by $C_\theta(x^\tau, \tau)$ where the superscript indicates the diffusion index (per the notation above). Given a frozen diffusion teacher $\Psi_\varphi$ (for instance a $K$-step DDIM teacher that follows the same subsampled schedule $\{\tau_k\}$, consistency distillation minimizes the squared regression objective

$$\mathcal{L}_{\mathrm{CD}}(\theta) = \mathbb{E}_{x_0,\tau,\varepsilon}\left[\left\|C_\theta(x^\tau, \tau) - \mathrm{sg}\left[\Psi_\varphi(x^\tau, \tau \to 0)\right]\right\|_2^2\right], \tag{S23}$$

where sg[·] denotes stop-gradient and $\Psi_\varphi(x^\tau, \tau \to 0)$ is the teacher's output after running the teacher's denoising chain from $x^\tau$ down to (approximate) $x^0$. The teacher must be run using the same subsampled schedule $\tau_k$ that the student will emulate or distill from. At inference time a consistency model requires only a single evaluation:

$$x^0 \approx C_\theta(x^{\tau_K}, \tau_K), x^{\tau_K} \sim \mathcal{N}(0, \mathbf{I}). \tag{S24}$$

## Diffusion Policy and RL Fine-tuning

Diffusion Policy (*3*) performs diffusion over robot actions conditioning on observations. Given an observation $o$, we roll out along the $K$-step subsampled schedule $\{\tau_K > \cdots > \tau_1\}$:

$$a^{\tau_{k-1}} = f_\theta(a^{\tau_k}, \tau_k | o), k = K, \ldots, 2, \tag{S25}$$

where $f_\theta$ follows the DDIM schedule as in Equation (S17) and (S18) conditioned on $o$. Then we can retrieve a clean action $a_t := a^{\tau_0}$ from the so-called diffusion policy.

**RL formulation.** Now we embed the $K$-step diffusion process as a sub-MDP into a single step in robotic manipulation MDP (*28*). Each denoising step can be viewed as sampling a cleaner noisy action $a^{\tau_{k-1}}$ from the Gaussian sub-policy in Equation (S20). We therefore model the process as a $K$-step sub-MDP with:

- **Initial state:** $s^K = (a^{\tau_K}, \tau_K, o)$ with $a^{\tau_K} \sim \mathcal{N}(0, \mathbf{I})$.

- **State:** $s^k = (a^{\tau_k}, \tau_k, o), k = K, \ldots, 1$.
- **Action:** $u^k = a^{\tau_{k-1}}$ drawn from the denoising sub-policy $\pi_\theta(u^k \mid s^k) = \mathcal{N}(\mu_\theta(a^{\tau_k}, \tau_k, o), \sigma^2_{\tau_k}\mathbf{I})$.
- **Transition:** $s^{k-1} = (u^k, \tau_{k-1}, o)$.
- **Reward:** this sub-MDP only receives terminal reward $R(a^{\tau_0})$ from the upper environment MDP.

The log-likelihood defined in Equation (S21) then computes the density of sub-policies, enabling end-to-end optimization of *task* rewards with respect to the denoising *sub-policies* via policy-gradient updates (e.g., PPO (*49*)).

**Handling single action vs. action chunk.**

Our framework supports both single-step and chunked action execution, which affects value computation and credit assignment:

- **Single-action:** Standard MDP formulation with per-step rewards $R_t$ and discount $\gamma$
- **Action-chunk:** Each chunk of $n_c$ actions is treated as a single decision. The chunk receives cumulative reward $R_{\text{chunk}} = R_{t:t+n_c-1} = \sum_{j=0}^{n_c-1} \gamma^j R_{t+j}$ , and the equivalent discount factor between chunks is $\gamma^{n_c}$.

For simplicity, we present the single-action case below; the chunked case works similarly by replacing per-step quantities with their chunk equivalents.

# SIMULATION RESULTS AND ANALYSIS

To complement our real-robot evaluations, we benchmark RL-100 in simulation and compare it against state-of-the-art diffusion/flow–based offline-to-online RL methods, including DPPO (*28*), ReinFlow (*63*), and DSRL (*52*). We also conduct ablations to verify the contribution of each component in RL-100.

## Main Results

We evaluate on three standard continuous-control suites that cover both dexterous manipulation and locomotion: Adroit (*64*), Meta-World (*65*), and MuJoCo locomotion (*66*, *67*). Unless noted otherwise, observations include robot proprioception together with visual inputs (RGB frames or point clouds), and actions are continuous joint/EE commands. We report normalized returns and task success rates (mean $\pm$ std over multiple seeds) and compare RL-100 with baselines finetuning diffusion/flow policies as well as their offline/online training protocols.

As shown in Fig. S6, RL-100, shown in red curves, consistently achieves the highest or near-highest performance across all ten tasks. On MuJoCo locomotion, RL-100 reaches ~10,000 return on halfcheetah-medium-v2, which is 2.2× higher than DPPO's 4,500 and 3.3× higher than DSRL's 3,000, and substantially outperforms all baselines on hopper and walker2d. On Adroit dexterous tasks, RL-100 attains near-perfect success rates, about 100%, on door and hammer, while DPPO plateaus at 0.9 and ReinFlow struggles below 0.6 on hammer. For Meta-World precision tasks like peg-insert-side, RL-100 maintains a stable 1.0 success rate after a few steps, where ReinFlow fails to exceed 0.2.

From a sample efficiency and stability perspective, RL-100 demonstrates faster convergence—achieving peak performance within 1-2M steps on most tasks versus 3-5M for

baselines—and exhibits notably lower variance (narrower confidence intervals) across random seeds. On challenging coordination tasks (pen, disassemble), RL-100's advantage is most pronounced, showing both higher asymptotic performance and more stable learning trajectories. The consistent gains across diverse modalities—from high-dimensional locomotion to contact-rich manipulation and precision insertion—validate RL-100's generality as a unified framework for diffusion policy fine-tuning.

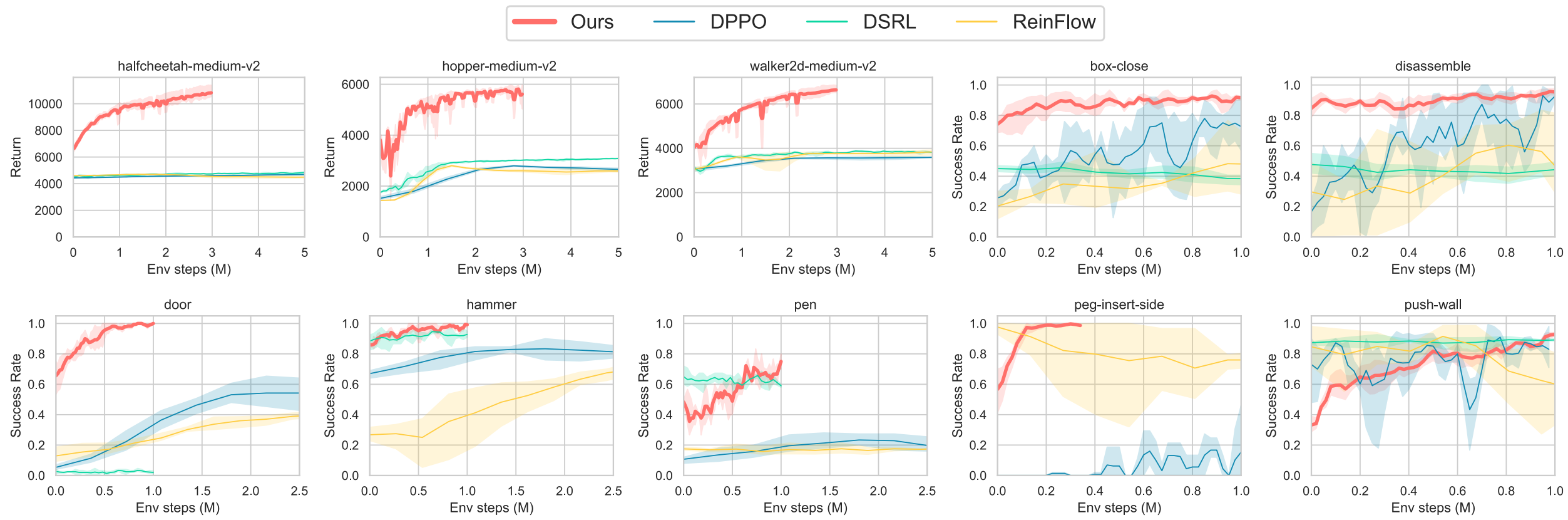

**Fig. S6. Learning curves of various methods on MuJoCo locomotion, Adroit, Meta-World** tasks are presented across three random seeds. Solid lines indicate the mean performance, while the shaded regions represent the corresponding 95% confidence interval. We use robot proprioception as the policy input for MuJoCo locomotion, and both 3D visual observations and proprioceptive state for Adroit (Door, Hammer, Pen) and Meta-World (Box-Close, Disassemble, Peg-Insert-Side, Push-Wall); baseline methods follow the input modalities specified in their original papers.

## Policy Inference Frequency

Fig. S7 presents a comparative analysis of our proposed methods, RL-100 (CM), against representative baselines in terms of model size (number of parameters) and policy inference speed (frequency in (Hz)). The results demonstrate the superior efficiency of our approach. Specifically, RL-100 (CM) with a Skip-Net architecture achieves an exceptional inference frequency of 378 (Hz), outperforming DSRL (35 (Hz)) and DPPO (30 (Hz)) by over an order of magnitude, while maintaining a compact model size of just 3.9M parameters *w.r.t.* DSRL (52.3M). Even a larger U-Net architecture (39.2M) enjoys an inference speed boost up to 133 (Hz) via CM. Reinflow benefits from a relatively high frequency due to its smaller network size, but it is still notably lower than ours, even with a smaller model.

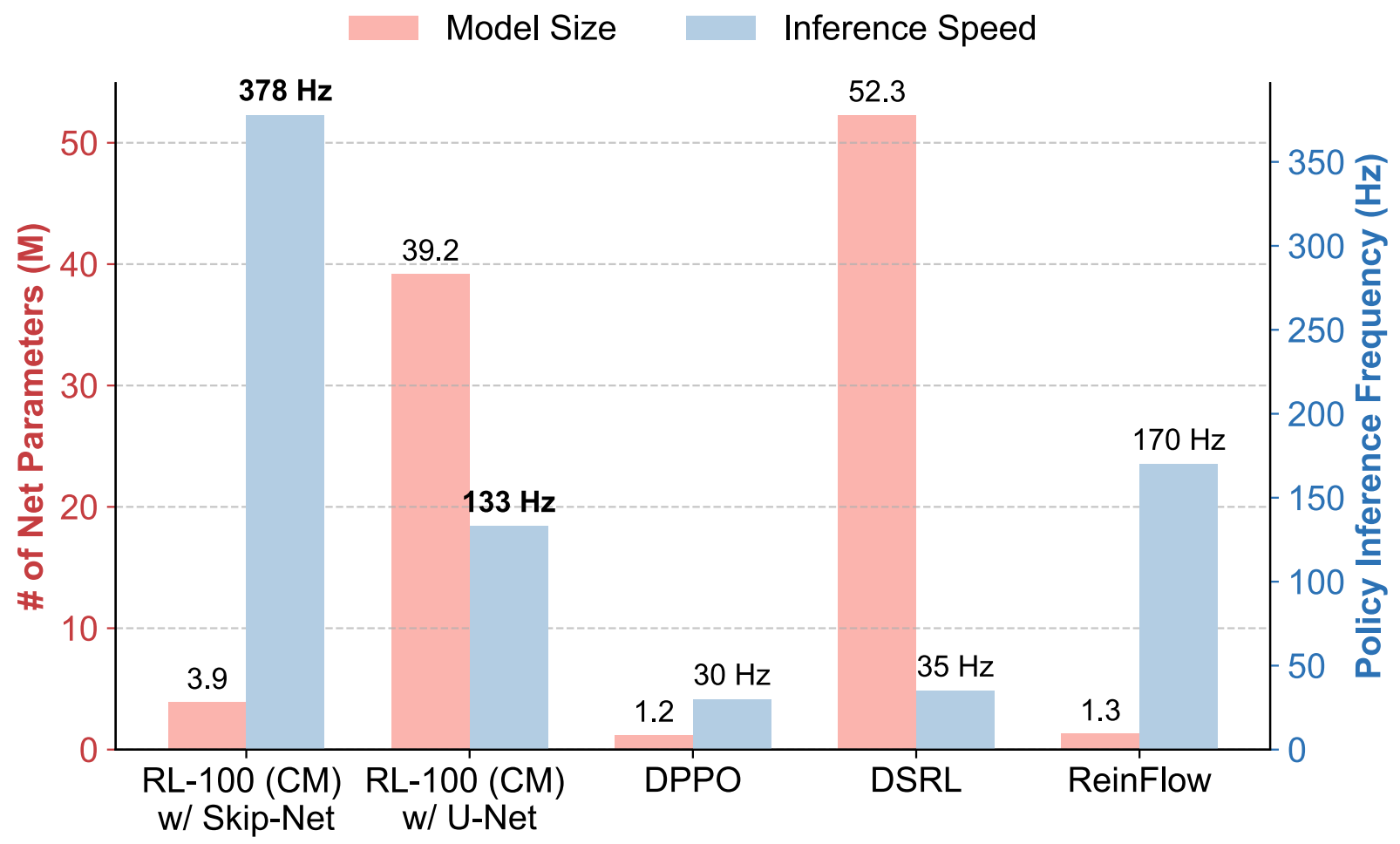

**Fig. S7. Comparison of two RL-100 variations with different action heads against other baselines in simulation in terms of model size and inference speed.** Model size is defined as the number of parameters

in a policy neural net, while inference speed is the frequency of a policy outputting an action given observations at each timestep.

This significant performance gain is attributed to our core methodology: we distill a diffusion policy into a consistency model (CM) capable of single-step inference alongside training/finetuning the original policy. Unlike diffusion-based policies that inherently rely on a slow, multi-step iterative sampling process, CM collapses policy generation into a single, efficient forward pass. Consequently, we achieve near real-time policy inference, effectively removing the decision-making model as a bottleneck. The overall system frequency becomes primarily limited by the speed of the perception system (e.g., camera frame rate at 30(Hz)), enabling truly reactive control and timely responses. This low latency is critical for robust performance in dynamic environments, where the system delay introduced by iterative samplers can severely hinder task success, which has been shown in our real-world experiments.

## Ablation Studies

We conduct ablation studies on five components: (i) visual modality (2D vs. 3D), (ii) diffusion-noise clipping, (iii) sampler/model (CM vs. DDIM), (iv) representation learning during fine-tuning (ReconVIB / no ReconVIB / frozen encoder), and (v) diffusion parameterization ($\epsilon$- vs. $x_0$-prediction). The comparison is listed in Fig. S8, and the detailed configurations and results are provided below.

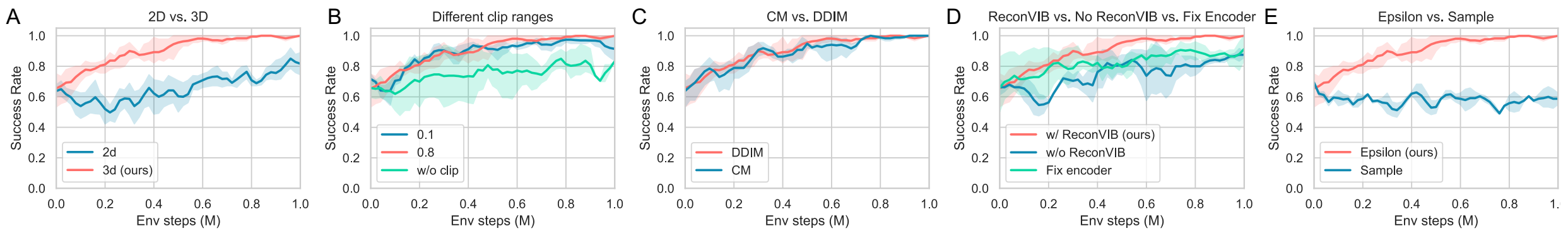

**Fig. S8. Learning curves of various ablation studies on the door task in Adroit are presented across three random seeds.** Solid lines indicate the mean performance, while the shaded regions represent the corresponding 95% confidence interval. From left to right: (A) 2D vision modality (i.e., images) vs. 3D point clouds used by RL-100. (B) Different clip ranges leveraged to bound the standard deviation of stochastic DDIM sampling steps. (C) Different models in RL-100. (D) Whether or not including reconstruction and VIB loss during finetuning as well as fixing encoder during finetuning. (E) Different diffusion parameterizations (i.e., the output of neural net; epsilon: predicting noises, sample: predicting clean samples.

**Visual modality: 2D vs. 3D.** Our framework is *modality-agnostic* and accepts either 2D RGB images or 3D point clouds as visual inputs. On the *Adroit door* task—a relatively clean scene—the 3D variant learns faster and attains a higher final success rate. We attribute this to the ability of 3D inputs to enable precise geometric filtering (e.g., ROI/radius ``crop") that cleanly isolates task-relevant structures such as the handle and contact surfaces, yielding a higher signal-to-noise ratio and easier credit assignment than 2D images. This ablation thus demonstrates both the flexibility of our framework (2D/3D optional) and the practical advantage of 3D in clean, contact-rich settings.

**Clipping the diffusion noise scale.** To ensure stable learning during RL fine-tuning, we introduce variance clipping in the stochastic DDIM sampling process. Specifically, we constrain the standard deviation at each denoising step:

$$\tilde{\sigma}_k = \text{clip}(\sigma_k, \sigma_{\min}, \sigma_{\max}),$$

where $\sigma_k$ is the original DDIM variance parameter from Equation(S18), and $[\sigma_{\min}, \sigma_{\max}]$ defines the permissible range. This modification effectively bounds the stochasticity of the behavior policy $\pi_\theta(a^{\tau_{k-1}}|a^{\tau_k}, \tau_k) = \mathcal{N}(\mu_\theta(a^{\tau_k}, \tau_k), \tilde{\sigma}_k^2\mathbf{I})$, preventing both:

- **Excessive exploration:** when $\sigma_t$ is too large, which can lead to out-of-distribution actions that

destabilize training or cause safety violations in physical systems.

- **Premature converge:** when $\sigma_t$ approaches zero, which eliminates exploration and prevents the policy from discovering better modes.

**Results.** As shown in Fig. S8, bounding the standard deviation of stochastic DDIM steps with a *moderate* clip (e.g., 0.8) yields the best stability–performance trade-off. An overly tight bound (0.1) under-explores and converges lower, whereas no clipping induces pronounced oscillations and wider intervals. We use 0.8 for Adroit, Mujoco Locomotion tasks, and real-world single-action mode tasks, and 0.1 for Meta-World and real-world chunk-action mode tasks.

**Consistency model (CM) vs. DDIM. One-step CM matches multi-step DDIM performance with $K\times$ speedup.** The distilled consistency model (CM, blue) achieves nearly identical learning curves and final success rates ($\sim$1.0) compared to the $K$-step DDIM policy (red), validating that our lightweight distillation successfully preserves task performance while enabling single-forward-pass inference. Both policies converge at similar rates and maintain comparable stability (overlapping confidence intervals), demonstrating that the CM effectively compresses the iterative denoising process without sacrificing control quality. This enables high-frequency deployment (100+ (Hz)) critical for dynamic real-world tasks.

**Representation learning during fine-tuning.** Including reconstruction plus VIB losses while updating the encoder achieves the highest and most stable success. Removing these losses degrades stability and final performance and freezing the encoder further limits gains; joint policy–representation adaptation mitigates representational drift and improves sample efficiency.

**Diffusion parameterization: $\boldsymbol{\epsilon}$ vs. $\mathbf{x_0}$.** Noise prediction ($\epsilon$-prediction) substantially outperforms clean-sample prediction ($x_0$-prediction), achieving ~40% higher final success ($\sim$1.0 vs.$\sim$0.6) with faster learning and reduced variance, suggesting better-conditioned gradients and closer alignment with the PPO objective used during RL post-training.

## Prediction Parameterization Analysis

Prior work on diffusion models for robot learning (e.g., DP3) predominantly adopts the $x_0$-prediction parameterization rather than $\epsilon$-prediction. The two parameterizations recover the clean action (or action chunk) $\hat{x}_0$ as

$$\hat{x}_0 = \frac{\left(x_t - \sqrt{1-\bar{\alpha}_t} f_\theta(x_t, t)\right)}{\sqrt{\bar{\alpha}_t}} \ (\epsilon\text{-prediction}),$$

$$\hat{x}_0 = f_\theta(x_t, t) \ (x_0\text{-prediction})$$

**Analysis.** A practical reason is variance amplification in $x_0$–prediction: early in the reverse process, i.e.,$t \to T$, the factor $1/\sqrt{\bar{\alpha}_t}$ becomes large, which magnifies estimation noise in $\varepsilon_\theta$ and yields a higher-variance $\hat{x}_0$. In supervised imitation settings (or when actions are executed open-loop), this extra variance can destabilize training and control, making the direct $x_0$-prediction attractive.

**Variance measurements.** We pretrain identical policies under the two parameterizations and, during environment rollouts, perform 100 forward passes per time step to estimate predictive variance. On a locomotion task (Hopper), the empirical variance of $\hat{x}_0$ is 0.1316 for $\epsilon$-prediction versus 0.0870 for $x_0$-prediction; on a manipulation task (Adroit–Door) it is 0.0589 **for $\boldsymbol{\epsilon}$-prediction versus** 0.0290 **for $\boldsymbol{x_0}$-prediction**, respectively.

**Implication for online RL.** In our *two-level MDP*—where each action is produced by a $K$-step denoising trajectory—the higher stochasticity induced by $\epsilon$–prediction acts as structured exploration *within* the denoising chain, improving coverage of latent action modes and helping avoid local optima during policy-gradient fine-tuning. With all other components fixed, online RL with $\epsilon$-prediction achieves faster and more reliable improvement on Adroit–Door (see Fig. S8), whereas $x_0$-prediction exhibits slower improvement and more frequent premature convergence.

**Takeaway.** While $x_0$-prediction can be preferable for low-variance imitation and open-loop execution, the variance amplification inherent to $\epsilon$-prediction is beneficial for *exploration* in diffusion-based online RL. Consequently, we adopt $\epsilon$-prediction for post-training, leveraging its exploratory behavior inside the denoising steps.

# SUPPLEMENTARY EXPERIMENT DETAILS

This section presents the setup and methodology for conducting real-world robot manipulation experiments. The experimental framework is designed to integrate perception, control, and demonstration collection in a coherent pipeline. We employ the Intel RealSense L515 camera to capture depth images, which are subsequently processed into 3D point clouds and transformed into the robot's root frame for consistent spatial representation. Intrinsic and extrinsic calibrations ensure accurate mapping between image coordinates and the robot workspace, while spatial filtering and downsampling produce compact yet informative point cloud observations suitable for high-frequency control.

The robotic platform consists of three manipulators—UR5, xArm, and Franka Emika Panda—equipped with dexterous or gripper-type end-effectors, as well as a passive fixed effector. Asynchronous control schemes are implemented across all manipulators to achieve low-latency, high-frequency execution, with task-specific interpolation strategies ensuring smooth and responsive motion.

Demonstrations for eight manipulation tasks are collected using teleoperation interfaces matched to task complexity. For dexterous 3D tasks, the Apple Vision Pro provides natural hand tracking, which is converted into robot joint or Cartesian commands via inverse kinematics for the LeapHand. For planar or low-dimensional tasks, a standard game joystick is used to provide delta-action control of the end-effector in the plane. Each task includes approximately 100 demonstration trajectories, efficiently recorded within a few hours per task.

Overall, the experimental setup provides a consistent, flexible, and practical framework for acquiring high-quality real-world data and evaluating robot manipulation policies across a diverse set of tasks.

## Camera Calibration

In the real-world experiments, we employ the Intel RealSense L515 camera, chosen for its ability to provide high-fidelity 3D point cloud observations owing to its superior depth accuracy. This ensures reliable perception for robot manipulation tasks across diverse environments.

The intrinsic calibration is performed using a Charuco board, which serves as the reference pattern for estimating the camera's internal parameters. Multiple images of the board are captured from varying angles and distances. The board corners are then detected and used to compute the intrinsic matrix and distortion coefficients through an optimization process that minimizes the reprojection error. This procedure guarantees an accurate mapping between 2D image coordinates and their corresponding 3D spatial points.

The extrinsic calibration aligns the camera frame with the robot's root frame using an AprilTag placed on the tabletop. First, a manual measurement establishes the transformation $T_{\text{tag2root}}$, which encodes the pose of the tag relative to the robot base. Subsequently, the AprilTag is detected in the camera's RGB channel, yielding the transformation $T_{\text{tag2camera}}$ from the tag to the camera frame. The camera-to-root transformation is then obtained as:

$$T_{\text{camera2root}} = T_{\text{tag2root}} \cdot (T_{\text{tag2camera}})^{-1}$$

This calibrated extrinsic transformation is stored and utilized throughout the experiments to maintain consistent spatial alignment between the camera observations and the robot workspace.

## Point Cloud Processing

The raw depth maps captured by the RealSense L515 are first back-projected into 3D point clouds using the calibrated intrinsic parameters, thereby establishing a metric reconstruction of the scene in the camera frame. These point clouds are then transformed into the robot's root frame using the calibrated extrinsic transformation $T_{\text{camera2root}}$, ensuring spatial consistency between visual observations and the robot's operational workspace.

To suppress environmental noise and spurious reflections, a spatial bounding box is applied in the root frame, filtering out irrelevant regions while preserving only the volume above the tabletop. This step enhances the semantic relevance of the retained points, focusing on task-relevant structures.

Finally, to improve computational efficiency while maintaining geometric fidelity, farthest point sampling (FPS) is applied. The processed point cloud is uniformly downsampled to 512 points, yielding a compact yet informative representation suitable for robot perception and policy learning. During deployment, the camera operates in a dedicated thread with a buffered pipeline, enabling asynchronous acquisition of observations. This design minimizes latency and allows the effective point cloud update rate to exceed 25 Hz, ensuring timely and reliable perception for closed-loop control.

## Robotic Arm & End-effector Control

The experimental platform incorporates three robotic manipulators—UR5, xArm, and Franka Emika Panda—equipped with different types of end-effectors, including the LeapHand dexterous hand, the Robotiq 2F-85 gripper, and a custom 3D-printed fixed effector. All control modules are designed to operate asynchronously, ensuring low-latency communication and high-frequency execution for responsive and stable manipulation.

For the UR5, relative action commands are transmitted at 30 Hz. These commands are handled through the Real-Time Data Exchange (RTDE) interface, where they are interpolated to 125 Hz for execution. A lookahead time of $1/30$ s and a control gain of 200 are used, providing a balance between smooth motion trajectories and timely responsiveness.

For the xArm, control is performed via absolute Cartesian position commands at 30 Hz. Linear interpolation is applied to the position trajectories, while orientation is represented in quaternion form and interpolated via spherical linear interpolation (slerp). The resulting trajectory is executed at 200 Hz, yielding continuous and smooth motion in task space.

For the Franka Emika Panda, absolute Cartesian position commands are also issued at 30 Hz. The robot's low-level controller performs trajectory interpolation, converting the input into control signals at 1000 Hz, thereby ensuring precise and compliant execution.

Regarding the end-effector control, both the LeapHand and the Robotiq 2F-85 are driven asynchronously at 30 Hz. The policy performs end-to-end joint control for the LeapHand, while the gripper uses binary control for the Soft-towel Folding task and continuous control for the Orange Juicing task.

The custom 3D-printed fixed effector, by contrast, is purely passive and requires no active control, serving as a simple but reliable interface for specific experimental settings like Dynamic Push-T or Agile Bowling.

## Data Collection

Data for eight real-world manipulation tasks are collected using teleoperation interfaces selected to match the complexity and dimensionality of each task. These tasks include both dexterous three-dimensional manipulations—such as folding, pouring, unscrewing, and juicing—and planar or constrained motions, such as Dynamic Push-T and Agile Bowling. The data collection strategy aims to provide sufficient coverage of the task space while maintaining practical efficiency in demonstration acquisition.

For tasks involving complex 3D hand motions, the Apple Vision Pro is employed. Its built-in spatial hand-tracking captures natural hand and finger movements, which are mapped to robot end-effector motions. Before starting each demonstration, the operator's hand pose is aligned with the robot end-effector through an initialization procedure, ensuring a consistent reference frame. During teleoperation, wrist and finger motions are streamed in real time and converted into robot joint or Cartesian commands, while gripper grasping actions are triggered via pinch gestures. For the LeapHand dexterous hand, inverse kinematics is solved in PyBullet to produce feasible joint configurations from the tracked hand poses. This setup allows operators to efficiently provide demonstrations for tasks requiring dexterous, spatially rich manipulations.

For planar or low-dimensional manipulation tasks—such as Dynamic Push-T and Agile Bowling—a standard game joystick is used. The joystick's XY-axis values are mapped to delta actions of the robot end-effector in the plane. This approach provides an intuitive and efficient means for recording demonstrations where precise two-dimensional control is sufficient.

Our dataset is intentionally varied. Some tasks feature a large number of demonstrations to test performance limits, while others use smaller datasets to evaluate sample efficiency.

# ADDITIONAL RELATED WORK

## Reinforcement Learning with Generative Diffusion Models

The integration of generative diffusion models with RL represents a paradigm shift in policy representation and optimization. Building upon foundational work in diffusion models (*14*, *15*) and flow matching (*16*), recent advances have demonstrated the power of these generative frameworks in capturing complex, multimodal action distributions inherent in RL problems. Diffusion Q-Learning (DQL) (*17*) pioneered this integration by replacing traditional Gaussian policies with conditional diffusion models in offline reinforcement learning, addressing fundamental limitations of parametric policies in modeling multimodal behaviors. This approach has evolved through multiple directions: weighted regression methods (*18–20*) train diffusion policies through importance-weighted objectives to maximize learned Q-functions; reparameterization gradient approaches (*21–23*) utilize gradient-based optimization despite temporal backpropagation challenges; and sampling-based methods (*24*) provide effective but computationally expensive solutions. More recently, consistency-based extensions (*25*) have further generalized diffusion and consistency policies to visual RL.

Recent works have also explored using RL to directly optimize diffusion models beyond traditional likelihood-based training. (*26*) demonstrate fine-tuning diffusion models with RL to maximize arbitrary reward functions, while (*27*) apply similar techniques specifically to text-to-image generation using human feedback. (*28*) introduce policy gradient methods tailored for diffusion-based policies, enabling effective online optimization while maintaining the expressiveness benefits of diffusion models.

To address computational bottlenecks inherent in diffusion models, (*29*) present Flow Q-Learning (FQL), which leverages flow-matching policies to model complex action distributions while avoiding recursive backpropagation through denoising processes. By training an independent single-step policy that matches the flow model's output, FQL achieves computational efficiency without sacrificing expressiveness, demonstrating state-of-the-art performance on offline RL benchmarks with significant computational advantages, particularly in offline-to-online fine-tuning settings. Similarly, One-Step Flow Q-Learning (OFQL) (*30*) extends this framework for even faster single-step action generation.

## Offline-to-Online Reinforcement Learning

The transition from offline pretraining to online fine-tuning presents unique challenges in managing distribution shift and preventing catastrophic forgetting while enabling continuous improvement. Conservative Q-Learning (CQL) (*44*) established the foundational framework for safe offline RL through pessimistic value estimation, preventing overestimation for out-of-distribution actions. Advantage-Weighted Regression (AWR) (*45*) provides a scalable framework that has influenced numerous subsequent works, though its restriction to Gaussian policies limits expressiveness for complex behavioral patterns. Calibrated Q-Learning (Cal-QL) (*46*) addresses initialization challenges by ensuring conservative Q-values are appropriately scaled for effective online fine-tuning, providing crucial insights for successful offline-to-online transfer. Uni-O4 (*47*) directly applies the PPO (*49*) objective to unify offline and online learning, eliminating the need for extra regularization.

Recent advances focus on efficiently leveraging both offline and online data through hybrid strategies. RLPD (*50*) achieves sample-efficient online learning by mixing offline and online experiences, demonstrating that careful data mixture strategies can accelerate learning significantly. (*51*) established fundamental frameworks for combining offline pretraining with online fine-tuning, showing that hybrid approaches achieve superior sample efficiency compared to pure online or offline methods. A paradigm shift in offline-to-online adaptation is represented by (*52*), which introduces DSRL (Diffusion Steering with Reinforcement Learning) that operates RL entirely in the latent noise space of pretrained diffusion policies rather than modifying base policy weights.